%% file: main.tex
\documentclass[10pt,twocolumn,letterpaper]{article}

\usepackage[pagenumbers]{cvpr} 

\input{preamble}
\definecolor{cvprblue}{rgb}{0.21,0.49,0.74}
\usepackage[pagebackref,breaklinks,colorlinks,allcolors=cvprblue]{hyperref}

\def\paperID{****} 
\def\confName{CVPR}
\def\confYear{2026}

\usepackage{multirow}
\usepackage{tikz}
\usepackage{colortbl}
\usepackage{makecell}
\newcommand{\hi}[1]{\textbf{\textcolor{other flat1}{#1}}}
\def\BibTeX{{\rm B\kern-.05em{\sc i\kern-.025em b}\kern-.08em
    T\kern-.1667em\lower.7ex\hbox{E}\kern-.125emX}}

\usepackage[table]{xcolor}
\usepackage{colortbl}
\usepackage{graphicx}

\definecolor{other flat1}{RGB}{175,0,75}
\definecolor{redbg}{RGB}{255,200,200}
\definecolor{greenbg}{RGB}{200,255,200}

\usepackage{pifont}
\newcommand{\cmark}{\ding{51}}
\newcommand{\xmark}{\ding{55}}

\usepackage{bbding}

\usepackage[accsupp]{axessibility}  

\title{UniPR: \underline{Uni}fied Object-level Real-to-Sim \underline{P}erception and \underline{R}econstruction \\ from a Single Stereo Pair}

\author{
    Chuanrui Zhang$^{1,2*}$ \quad
    Yingshuang Zou$^{4*}$ \quad
    Zhengxian Wu$^{5*}$ \quad
    Yonggen Ling$^{2,3\dagger}\textsuperscript{\Envelope}$ \\
    Yuxiao Yang$^{5}$ \quad
    Ziwei Wang$^{1}\textsuperscript{\Envelope}$ \\
    \vspace{0.5cm}
    $^{1}$NTU \qquad $^{2}$Tencent Robotics X \qquad $^{3}$Futian Laboratory \qquad $^{4}$HKUST \qquad $^{5}$THU
}

\makeatletter
\def\blfootnote{\gdef\@thefnmark{}\@footnotetext}
\makeatother

\begin{document}

\twocolumn[{%
\renewcommand\twocolumn[1][]{#1}%
\maketitle

\vspace{-1.0cm} 
\begin{center}
    \large \url{https://xingyoujun.github.io/unipr}
\end{center}
\vspace{0.1cm} 

\vspace{-0.4cm}
\begin{center}
    \centering
    \captionsetup{type=figure}
    \includegraphics[width=0.95\textwidth]{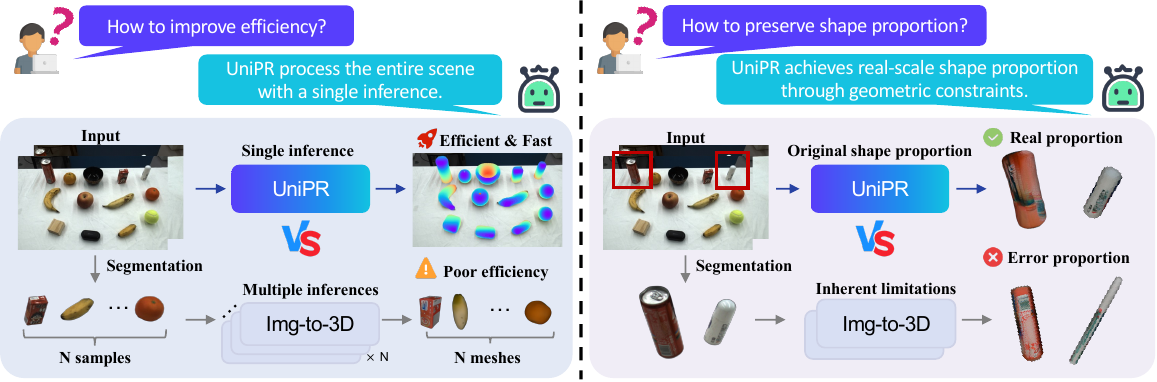}
    \vspace{-0.2cm}
    \captionof{figure}{
    \textbf{UniPR} enables efficient full-scene processing and produces real-scale, physically consistent object shapes by leveraging geometric constraints, outperforming traditional image-to-3D reconstruction models. 
    As the first fully end-to-end method, UniPR achieves up to 100× faster generation and delivers 3x improvements in shape-proportion accuracy compared with SOTA mehtods.
    }
    \vspace{-0.1cm}
    \label{fig:teaser}
\end{center}%
}]

\blfootnote{$^*$Equal contribution. $^\dagger$Project leader. $\textsuperscript{\Envelope}$Corresponding authors. \\ 
Work was done at Tencent Robotics X.
}

\input{sec/0_abstract}    
\input{sec/1_intro}

\input{sec/2_related}
\input{sec/3_method}
\input{sec/4_exp}
\input{sec/5_con}

\section*{Acknowledgments}
This work was supported by the Singapore National Robotics Programme Research Project under Grant DS-RFM M25N4N2009.

{
    \small
    \bibliographystyle{ieeenat_fullname}
    \bibliography{main}
}
\input{sec/X_suppl}

\end{document}

%% file: sec/0_abstract.tex
\vspace{-0.6cm}
\begin{abstract}
Perceiving and reconstructing objects from images are critical for real-to-sim transfer tasks, which are widely used in the robotics community.
Existing methods rely on multiple submodules such as detection, segmentation, shape reconstruction, and pose estimation to complete the pipeline.
However, such modular pipelines suffer from inefficiency and cumulative error, as each stage operates on only partial or locally refined information while discarding global context.
To address these limitations, we propose \textbf{UniPR}, the first end-to-end object-level real-to-sim perception and reconstruction framework.
Operating directly on a single stereo image pair, UniPR leverages geometric constraints to resolve the scale ambiguity.
We introduce Pose-Aware Shape Representation to eliminate the need for per-category canonical definitions and to bridge the gap between reconstruction and pose estimation tasks.
Furthermore, we construct a large-vocabulary stereo dataset, LVS6D, comprising over 6,300 objects, to facilitate large-scale research in this area.
Extensive experiments demonstrate that UniPR reconstructs all objects in a scene in parallel within a single forward pass, achieving significant efficiency gains and preserves true physical proportions across diverse object types, highlighting its potential for practical robotic applications.

\end{abstract}

%% file: sec/1_intro.tex
\vspace{-0.6cm}
\section{Introduction}
\label{sec:intro}
\vspace{-0.1cm}

\begin{figure*}[tb]
  \centering
  \vspace{-0.5cm}
  \includegraphics[width=1\textwidth]{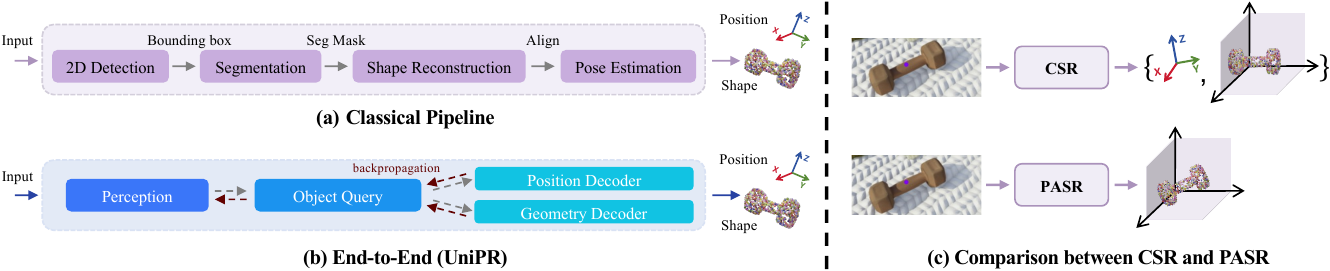}
  \vspace{-0.4cm}
  \caption{
  Comparison between our end-to-end approach and the classical pipeline.
Our method enables information to flow seamlessly across all components, allowing the network to leverage the full-image context for shape reconstruction. 
This end-to-end design effectively handles occlusion and significantly improves the preservation of true shape proportions compared to classical, modular pipelines.
  }
  \label{fig:e2e}
  \vspace{-0.5cm}
\end{figure*}


Accurate perception and reconstruction of real-world objects are fundamental to robot manipulation~\cite{jiang2022ditto, huang2024rekep, zhang2025imowm} and scene understanding~\cite{jia2025sceneverse, li2023voxelformer, zhang2025transplat, zou2025mudg, zou2024m}.
These applications require not only visual fidelity but also physical accuracy in object geometry and position to ensure reliable interaction between real and simulator.
However, existing approaches typically decompose this process into isolated modules, such as detection, segmentation, shape reconstruction and pose estimation, which suffer from error propagation, computational inefficiency, and inconsistent shape proportion across the pipeline.
This work aims to bridge this critical gap by proposing a unified framework that directly transforms raw sensor inputs into physically accurate 3D reconstructions, preserving true object proportions and spatial relationships for seamless real-to-sim transfer.

Current approaches to object-level real-to-sim reconstruction can be broadly categorized into three types, each exhibiting fundamental limitations in achieving scalable physical accuracy.
Instance-level methods~\cite{xiang2017posecnn, tekin2018real, lee2025any6d} rely on CAD models or multi-view images to reconstruct objects using NeRF\cite{mildenhall2021nerf} and subsequently predict the relative pose with respect to the corresponding object model.
However, such reliance on object-specific priors severely constrains their generalization ability to unseen objects, particularly in the absence of multi-view scans.
Category-level approaches~\cite{wang2019nocs, sun2022onepose, he2022onepose++} reconstruct object shapes within predefined canonical spaces while simultaneously predicting relative poses.
However, their reliance on category-specific shape priors severely limits practical scalability as most methods handle fewer than six categories because defining canonical spaces and accommodating large intra-class variations is inherently challenging.

Recently, large-scale 3D generative models~\cite{xiang2025structured, hunyuan3d2025hunyuan3d} have demonstrated remarkable zero-shot capability and high visual fidelity.
Combining instance-level methods with object geometries reconstructed from large image-to-3D models~\cite{geng2025one} has emerged as a promising direction for real-to-sim transfer of unseen objects.
However, image-to-3D methods often struggle to preserve shape proportion, as they generate object meshes solely from visual inputs lacking metric information.
Furthermore, these methods can typically handle only a single object at a time, limiting their applicability to complex scenes containing multiple objects, as shown in \Cref{fig:teaser}.
Crucially, most existing pipelines require predefined object bounding boxes and segmentation masks, leading to error propagation and computational inefficiency.

Furthermore, monocular approaches~\cite{geng2025one, he2022onepose++} inherently suffer from scale ambiguity, making it impossible to obtain the metric-accurate reconstructions required for robotic manipulation. 
In contrast, stereo vision provides the geometric constraints necessary to resolve this ambiguity and preserve object shape proportions, forming a robust foundation for reliable real-to-sim transfer that maintains both geometric fidelity and metric accuracy~\cite{liu2020keypose,chen2023stereopose,zhang2024category}.

To overcome the limitations of existing approaches, we propose a unified end-to-end framework that seamlessly integrates perception and reconstruction for real-to-simulation transfer, as illustrated in \Cref{fig:e2e}(ab).
At the core of this framework is the Pose-Aware Shape Representation (PASR), which jointly encodes object pose and geometry directly in the observation space. This design removes the dependence on predefined canonical spaces and enables tight coupling between pose estimation and shape reconstruction, as shown in \Cref{fig:e2e}(c).
This representation enables the direct prediction of physically accurate 3D shapes aligned with true world coordinates, effectively removing the error-prone decoupling between pose estimation and shape reconstruction.
Building upon this representation, we introduce \textbf{UniPR}, a single-forward pipeline that processes raw stereo images to simultaneously detect multiple objects and reconstruct their 3D geometries in parallel.
Unlike traditional sequential pipelines that handle objects individually after detection and segmentation, our framework processes all scene objects concurrently through object queries in a transformer decoder, thereby achieving substantial improvements in computational efficiency and mitigating error accumulation.
The stereo-based design provides essential geometric cues that resolve the scale ambiguities inherent in monocular methods, ensuring that reconstructed objects maintain accurate positions and shape proportions.

To validate our approach, we construct \textbf{LVS6D}, a large-scale stereo dataset encompassing over 6,300 diverse objects, which demonstrates the scalability of our representation without the need for predefined canonical spaces.
Comprehensive experiments demonstrate UniPR achieves superior shape proportion accuracy compared to generative 3D models such as HunYuan3D~\cite{hunyuan3d2025hunyuan3d} and Trellis~\cite{xiang2025structured}, even when provided with identical 2D masks and poses.
Furthermore, our framework efficiently processes multiple objects in parallel, achieving up to a \textbf{100×} speed-up, while exhibiting robust generalization to real-world scenes and unseen object categories.
These capabilities enable direct deployment in robotic manipulation tasks that demand precise physical alignment between real and reconstructed objects, as confirmed through extensive real-world experiments.

In summary, our main contributions are as follows:
\begin{itemize}
\item[$\bullet$] We introduce the first end-to-end real-to-sim perception-reconstruction framework, eliminating intermediate modules and preventing error propagation.
\item[$\bullet$] We propose PASR, which enables seamless scaling to hundreds of object categories while preserving true physical proportions and spatial relationships.
\item[$\bullet$] Our framework supports parallel batch processing of multiple objects in a single forward pass, substantially improving computational efficiency.
\item[$\bullet$] We construct LVS6D, a large-scale stereo dataset with diverse object geometries for real-to-sim transfer tasks.
\end{itemize}

%% file: sec/2_related.tex
\vspace{-0.1cm}
\section{Related Work}
\vspace{-0.1cm}
\label{sec:related}

\subsection{3D Object Reconstruction}
\vspace{-0.1cm}
Current 3D reconstruction methods continue to face challenges in achieving both physical accuracy and computational efficiency.
Stereo and multi-view approaches~\cite{lee2025any6d, zhang2024category, irshad2022shapo, irshad2022centersnap} can recover metric scale by exploiting geometric principles such as triangulation from calibrated camera pairs.
However, these methods typically require multi-view images for NeRF-based~\cite{mildenhall2021nerf} reconstruction or rely on category-level shape priors as generative templates, which limits their efficiency in real-to-simulation transfer and hinders generalization to novel objects.
Recently, large-scale generative models~\cite{xiang2025structured, hunyuan3d2025hunyuan3d, yang2025wonder3d++, yang2025nova3d} have demonstrated remarkable zero-shot visual fidelity.
These methods leverage voxel- or vector-set–based latent representations to model object geometry and employ diffusion transformers~\cite{peebles2023scalable} conditioned on images to generate novel 3D instances.
However, they inherit the intrinsic scale ambiguity of monocular inputs and exhibit limited accuracy in preserving object shape proportions.
Moreover, these models typically process only a single object at a time and depend on upstream modules such as segmentation masks, which reintroduce error propagation and restrict their applicability to complex scenes.
In contrast, our framework introduces a unified end-to-end pipeline that integrates stereo geometric constraints to simultaneously detect and reconstruct multiple physically accurate objects, ensuring metric precision essential for real-world interaction.

\vspace{-0.1cm}
\subsection{Object-level Real-to-sim Transfer}\label{subsec:real2sim}
\vspace{-0.1cm}
Object-level real-to-simulation transfer aims to bridge the physical and virtual worlds by accurately reconstructing real-world objects with precise geometry, position, and scale for direct deployment in simulation environments~\cite{jiang2022ditto, huang2024rekep, he2022onepose++}.
Early approaches relied on instance-level reconstruction using accurate CAD models or multi-view scans~\cite{tekin2018real, lee2025any6d, zhang2024category, wen2024foundationpose}, with object poses estimated relative to predefined models.
To enhance generalization, the field advanced toward category-level understanding through frameworks such as the Normalized Object Coordinate Space (NOCS)~\cite{wang2019nocs}, which established a unified canonical space for objects within the same category.
Recent methods~\cite{tian2020shape, irshad2022centersnap, zhang2024category} further extended this paradigm by modeling deformations from category-specific shape priors, enabling joint pose and shape estimation for novel instances within known categories.
However, these approaches remain constrained by their reliance on predefined canonical spaces, limiting practical scalability to fewer than six categories due to the challenges of defining canonical orientations and handling intra-class variations~\cite{zhang2025omni6d}.
Recent advances explore zero-shot novel object reconstruction. 
Methods like Any6D~\cite{lee2025any6d} achieve impressive single-object reconstruction without category constraints, while OnePoseViaGen~\cite{geng2025one} leverages large-scale image-to-3D generative models to reconstruct object geometry before estimating pose.
Despite their visual quality, these methods typically process objects sequentially after 2D detection and segmentation, suffer from scale ambiguity in monocular settings, and fail to preserve true object proportions, resulting in limited physical accuracy.
In contrast to these fragmented pipelines, our work introduces the first unified end-to-end framework that simultaneously detects multiple objects and reconstructs their physically accurate 3D shapes.

\vspace{-0.1cm}
\subsection{Datasets for Object-level Real-to-sim Transfer}
\vspace{-0.1cm}
The NOCS dataset~\cite{wang2019nocs}, which encompasses the CAMERA and REAL datasets, is the most widely utilized for category-level object pose estimation and shape reconstruction tasks. 
CAMERA includes 6 categories with 1085 object instances, and REAL, mirroring CAMERA's categories, contains 42 instances. 
The Wild6D dataset~\cite{fucategory} comprises 5166 videos and 1.1 million images, spanning 1722 object instances across 5 categories. 
However, the limited category range in these datasets restricts their real-world generalization capabilities~\cite{fucategory, yen2021inerf}.
Building on these, the Omni6D dataset~\cite{zhang2025omni6d} offers sophisticated instance-level rotational invariance annotations for a wider range of 166 categories and 4688 instances. 
However, the process of annotating such rotation invariance is arduous and the datasets grapple with ambiguities in the alignment of functional and geometric properties, impeding their expansion to real-world contexts.
Contrastingly, our method presents an innovative annotation that integrates object shape and pose, sidelining the requirement for predefined category-level canonical spaces and greatly easing dataset expansion. 
Notably, our dataset readily encompasses 192 categories and over 6,300 instances.


%% file: sec/3_method.tex
\begin{figure*}[tb]
  \centering
  \vspace{-0.3cm}
  \includegraphics[width=0.87\textwidth]{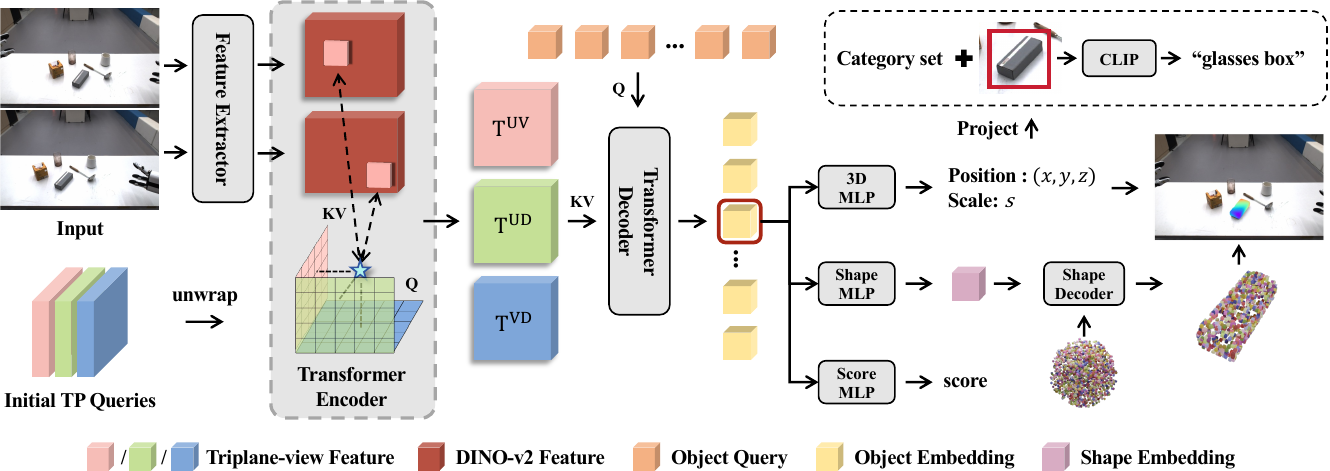}
  \vspace{-0.2cm}
  \caption{
  \textbf{Overview of Our Proposed UniPR.} 
We present UniPR, a single-forward network capable of simultaneously processing multiple unknown objects.
Taking stereo image pairs as input, UniPR first encodes the scene into Tri-Plane View features that comprehensively capture spatial and geometric information.
Within the transformer decoder, object queries are employed to extract instance-specific features from these TPV embeddings, enabling the network to reason about multiple objects in parallel.
The resulting object embeddings are then fed into specialized prediction heads to infer each object’s semantic label, 3D position, physical scale, and pose-aware shape representation.
  }
  \label{fig:pipeline}
  \vspace{-0.4cm}
\end{figure*}

\vspace{-0.2cm}
\section{Method}
\label{sec:method}
\vspace{-0.1cm}
We introduce a single-forward network named \textbf{UniPR}, that infers object position, size and pose-aware 3D shape from stereo images. 
An overview of our method is illustrated in \cref{fig:pipeline}. 
Formally, given stereo images as input, UniPR extracts 2D stereo features using DINOv2~\cite{oquab2023dinov2}. 
We then utilize a triplane-view representation as the global spatial framework to effectively aggregate stereo information. 
Next, the transformer decoder generates expressive object embeddings, which are subsequently used to predict the object's position, scale, and shape embeddings. 
Finally, the shape decoder reconstructs the object's pose-aware 3D shape.
To obtain the shape embedding and decoder, we first train a pose-aware shape Variational Auto Encoder (VAE).
Our VAE encodes rotated objects into compact object shape embeddings and decodes them into object occupancy. 
The pretrained VAE decoder is then integrated into the UniPR pipeline as the shape decoder.

\vspace{-0.1cm}
\subsection{Pose-Aware Shape VAE}
\label{sec:method_pasa}
\vspace{-0.1cm}
We first introduce our proposed pose-aware shape VAE. 
Unlike previous shape VAEs~\cite{kingma2013auto, zhang20233dshape2vecset}, our approach focuses on rotated objects and utilizes a more lightweight object embedding, which is critical for our detection pipeline. 
Traditional cubic voxel space encounters challenges when objects are rotated, as objects normalized to a unit cube may extend beyond this boundary upon rotation. 
Re-normalizing rotated objects within a cubic voxel space can lead to scale ambiguities, where the perceived scale varies with the object's rotation. To address this limitation, we introduce a \textbf{spherical voxel space} that normalizes objects using a unit sphere. This approach ensures that objects remain within the defined boundary regardless of their rotation, effectively preventing ambiguities in shape representation.

Given the surface point cloud $\mathbf{P}_{\rm surface} \in \mathbb{R}^{N \times 3}$ of object, we first apply positional embeddings to generate surface embeddings $\boldsymbol{z}_{\rm surface} \in \mathbb{R}^{N \times C}$.
Since these surface embeddings are high-dimensional and cannot be directly integrated into the detection pipeline, we aim to compress them to an object-level representation.
We initialize an object embedding $\boldsymbol{z}_{\rm object} \in \mathbb{R}^{C}$ and employ cross-attention layers to learn a mapping from surface to object embeddings.

\vspace{-10pt}
\begin{equation}
\boldsymbol{z}_{\rm object} = {\rm CrossAttn}(\boldsymbol{z}_{\rm object}, \boldsymbol{z}_{\rm surface})
\end{equation}
Next, we utilize two multi-layer perceptrons (MLPs) to predict the mean $\mu \in \mathbb{R}^{C_{\rm kl}}$ and variance $\sigma^2 \in \mathbb{R}^{C_{\rm kl}}$, forming a Gaussian distribution. 
We then apply KL regularization to ensure effective representation and project the embeddings into a lower-dimensional latent space.

For the decoder $\mathcal{D}_{\rm VAE}$ , we aim to compute the occupancy probability for each query point $\mathbf{P}_{\rm query} \in \mathbb{R}^{N \times 3}$ in order to recover the object shape. 
Firstly, We randomly sample an object latent embedding $\boldsymbol{z}_{\rm sampled} \in \mathbb{R}^{C_{\rm kl}}$ from the Gaussian distribution $(\mu, \sigma^2)$ obtained from the encoder.

\vspace{-10pt}
\begin{equation}
\boldsymbol{z}_{\rm sampled} = \mu + \sigma \cdot \epsilon
\end{equation}
where $\epsilon \sim \mathcal{N}(0,1)$.
We then use a cross-attention mechanism to transform the sampled embedding $\boldsymbol{z}_{\rm sampled}$ into surface embeddings $\hat{\boldsymbol{z}}_{\rm surface} \in \mathbb{R}^{N \times C}$ to recover the original surface distribution. 
To further enhance the representation ability, we apply several self-attention blocks to enhance the point embeddings. 
We then utilize positional embeddings to generate query embeddings $\boldsymbol{z}_{\rm query}$ and use cross-attention to map the surface distribution to the query embeddings.

\vspace{-10pt}
\begin{equation}
\boldsymbol{z}_{\rm query} = {\rm CrossAttn}(\boldsymbol{z}_{\rm query}, \hat{\boldsymbol{z}}_{\rm points})
\end{equation}
Finally, we employ an MLP to project the query point embeddings into occupancy values.

\vspace{-10pt}
\begin{equation}
\mathcal{O}(x,y,z) = \phi(\boldsymbol{z}_{\rm query}(x,y,z))
\end{equation}
where $\mathcal{O}(x,y,z)$ denotes the occupancy values of point $(x,y,z)$ in the spherical voxel space and $\phi$ is MLP layers.

\subsection{Triplane Encoder}
\label{sec:method_encoder}
In this section, we introduce the information encoder module of our proposed UniPR.
We employ a triplane mechanism~\cite{huang2023tri} to aggregate stereo features into a global coordinate system. Specifically, we initialize the Triplane-View (TPV) feature representation
$\mathbf{T} = [\mathbf{T}^{\rm UV}, \mathbf{T}^{\rm UD}, \mathbf{T}^{\rm VD}]$, where 
$\mathbf{T}^{\rm UV} \in \mathbb{R}^{U \times V \times C},
\mathbf{T}^{\rm UD} \in \mathbb{R}^{U \times D \times C}, 
\mathbf{T}^{\rm VD} \in \mathbb{R}^{V \times D \times C} $. 
Here, U and V denote the image plane dimensions, while D represents depth dimension. 
We construct the UVD space, which is better suited for object detection in camera space, facilitating accurate object representation and matching within the global coordinate framework.

Given the initialized TPV, we apply stereo cross-attention to lift stereo features onto TPV planes. 
For a voxel grid located at $(u,v,d)$ in UVD space, we aim to update the feature $\mathbf{T}(u,v,d)$ by separately refining $\mathbf{T}^{\rm UV}(u,v)$, $\mathbf{T}^{\rm UD}(u,d)$, $\mathbf{T}^{\rm VD}(v,d)$. 
To achieve this, we first backproject each voxel to determine the corresponding pixel in the stereo images. 
Leveraging the UVD space, we first obtain the pixel coordinates, after which we update the TPV features using stereo cross-attention, as described in \cref{eq:sca}.


\vspace{-20pt}
\begin{equation}
\begin{aligned}
& \mathbf{T}(u,v,d) = \mathcal{F}(\mathbf{T}(u,v,d), \mathbf{F}_l(u_{l}, v_{l}), \mathbf{F}_r(u_{r}, v_{r})) \\
\end{aligned}
\label{eq:sca}
\end{equation}
where $\mathbf{F}_l, \mathbf{F}_r \in \mathbb{R}^{H \times W \times C}$ denote the features extracted from the left and right stereo images, respectively. The function $\mathcal{F}$ represents stereo cross-attention which generates attention weights based on the TPV features to aggregate stereo image features effectively. 
Additionally, we apply self-attention to fuse information across the three TPV planes, enhancing the overall TPV representation. 
This combination of self-attention and cross-attention modules is repeated $N$ times, corresponding to $N$ encoder layers.

\vspace{-0.1cm}
\subsection{Object Embedding Decoder}
\vspace{-0.1cm}
\label{sec:method_decoder} 
We aim to generate object embeddings that encapsulate all relevant information about the corresponding objects, including position, scale, and 3D shape, and then decode these information from the object embeddings as the network's output. 
To achieve this, we adopt the standard transformer decoder structure from DETR~\cite{carion2020end}, which consists of $L$ decoder layers. 
Each layer includes three key components: a self-attention module to facilitate interactions among object queries, a cross-attention module to integrate stereo-aware image features, and a feed-forward network (FFN) to update the object queries. 
We employ multi-head attention to perform these attention operations effectively.

Through iterative interactions across the decoder layers, the model generates object embeddings that capture high-level representations. 
We present a 3D MLP to regress the object's position$(x,y,z)$ and scale$s$, and a shape MLP to generate the shape embedding distribution$(\mu,\sigma^2)$.
Here, the shape embedding distribution follows a KL-regularized Gaussian distribution, as discussed in \cref{sec:method_pasa}.
Finally, we obtain the object point cloud prediction by calculating the occupancy of spherical query points through the pretrained shape decoder. 
Combining position, scale, and 3D shape, we can retrieve the object shape in camera coordinates, which can further guide robot manipulation tasks.

Additionally, we eliminate the classification head used in previous work~\cite{zhang2024category}, as we find that the classification head may degrade object detection results, especially for categories that are difficult to distinguish. 
Instead, we utilize the CLIP model~\cite{radford2021learning} to determine the correct category from the preset categories using the 2D bounding box, which is the 2D projection of the 3D location result.
During inference, we regress the object confidence score to determine whether the object query corresponds to a valid object.


\subsection{Training Loss}
\label{sec:method_loss}
\vspace{-0.1cm}
To train the pose-aware shape VAE, we optimize the Binary Cross Entropy (BCE) loss for the occupancy prediction of the input query points, following prior works~\cite{zhang20233dshape2vecset, mescheder2019occupancy}:
\begin{equation}
\mathcal{L}_{\rm recon} = {\rm BCE}(\hat{\mathcal{O(\mathbf{X})}}, \mathcal{O(\mathbf{X})})
\end{equation}
where $\mathbf{X}$ represents the input query points, and $\hat{\mathcal{O}}, \mathcal{O}$ denote the predicted and ground truth occupancy values of these points. 
We also employ KL regularization as the KL loss with respect to the standard Gaussian distribution:
\vspace{-0.2cm}
\begin{equation}
\mathcal{L}_{\rm kl-reg} = \frac{1}{C_{\rm kl}}\sum_{j=1}^{C_{kl}}\frac{1}{2}(\hat\mu^2 + \hat\sigma^2 - log\hat\sigma^2)
\label{eq:kl-reg}
\vspace{-0.15cm}
\end{equation}
where $(\hat\mu, \hat\sigma^2)$ denotes the encoded shape distribution.

In the main detection pipeline, UniPR supervises the learning of object position$(x,y,z)$, scale$s$, and shape embedding distribution$(\mu,\sigma^2)$. 
Similar to the approach used in DETR~\cite{carion2020end}, we apply the Hungarian algorithm~\cite{kuhn1955hungarian} for one-to-one matching between ground truth and predicted values.
For the position and scale supervision, we employ $\mathcal{L}_1$ loss to measure the discrepancies.
We utilize KL regularization to quantify the distance between the ground truth shape distribution and the predicted distribution, as detailed in \cref{eq:kl}.
\begin{equation}
\mathcal{L}_{kl} = \frac{1}{C_{kl}}\sum_{j=1}^{C_{kl}}\frac{1}{2}(\frac{(\hat\mu-\mu)^2 + \hat\sigma^2}{\sigma^2} - log\hat\sigma^2 + log\sigma^2)
\label{eq:kl}
\end{equation}
where $(\hat\mu, \hat\sigma^2)$ denotes the predicted distribution and $(\mu, \sigma^2)$
is the ground truth.
The loss for detection pipeline can be summarized as follows:
\vspace{-0.1cm}
\begin{equation}
\mathcal{L}_{detection} = \mathcal{L}_{position} + \mathcal{L}_{scale} +  \lambda_{shape} \times \mathcal{L}_{shape} 
\end{equation}
where $\lambda_{shape}$ is a hyperparameter used to balance the contribution of the shape loss in the overall loss function.

%% file: sec/4_exp.tex
\begin{table*}[!t]
\centering
\vspace{-0.5cm}
\resizebox{0.80\linewidth}{!}{
\begin{tabular}{l|ccc|c|c|c|c|cc}
\toprule
\multirow{2}{*}{\textbf{Method} }&\multicolumn{3}{c|}{\textbf{Requirement}} & \multirow{2}{*}{\textbf{CD $\downarrow$ } } & \multirow{2}{*}{\textbf{CD@Top10 $\downarrow$}} & \multirow{2}{*}{\textbf{F-Score $\uparrow$}} & \multirow{2}{*}{\textbf{SPE $\downarrow$}} & \multicolumn{2}{c}{\textbf{Inference Time (s)}}  \\

& 2D & Seg & Pose & & & & & Single & Whole Image \\
\midrule
Trellis~\cite{xiang2025structured} & \cellcolor{redbg}\cmark & \cellcolor{redbg}\cmark & \cellcolor{redbg}\cmark & 0.1096 & 0.0220 & 0.334 & 0.475 & 8.62 & \cellcolor{redbg}43.08 \\
Hunyuan2.1~\cite{hunyuan3d2025hunyuan3d} & \cellcolor{redbg}\cmark & \cellcolor{redbg}\cmark & \cellcolor{redbg}\cmark  & 0.0644 & 0.0072 & 0.553 & 0.320 & 74.16 & \cellcolor{redbg}370.78 \\
Ours & \cellcolor{greenbg}\xmark & \cellcolor{greenbg}\xmark & \cellcolor{greenbg}\xmark  & \textbf{0.0083} & \textbf{0.0026} & \textbf{0.883} & \textbf{0.109} & \textbf{0.63} & \cellcolor{greenbg}\textbf{0.63} \\
\bottomrule
\end{tabular}
}
\vspace{-0.1cm}
\caption{
\textbf{Quantitative comparison on reconstruction.}
Both Trellis and HunYuan2.1 require perfect 2D bounding boxes, segmentation masks, and poses as inputs, whereas our method operates in an end-to-end manner directly from stereo images.
Here, CD denotes the Chamfer Distance, SPE refers to the Shape Proportion Error, and inference time is measured in seconds per object and per scene.
}
\label{tab:gen_result}
\vspace{-0.3cm}
\end{table*}

\begin{table*}[!t]
\centering
\vspace{-0.05cm}
\resizebox{0.70\linewidth}{!}{
\begin{tabular}{l|ccc|ccc|ccc}
\toprule
\multirow{2}{*}{\textbf{Method} }&\multicolumn{3}{c|}{\textbf{Easy}} &\multicolumn{3}{c|}{\textbf{Medium}}   &\multicolumn{3}{c}{\textbf{Hard}}  \\
     & AP$\uparrow$ & APE$\downarrow$ & ACD$\downarrow$ & AP$\uparrow$ & APE$\downarrow$ & ACD$\downarrow$ & AP$\uparrow$ & APE$\downarrow$ & ACD$\downarrow$     \\
\midrule
Coders~\cite{zhang2024category}   & 0.102   & 2.200  & 2.348 & 0.112 & 2.144 & 3.084 & 0.070 & 2.230 & 10.146 \\
Coders with PASR  & 0.495   & 1.332  & 0.850 & 0.572 & 1.522 & 1.225 & 0.483 & 1.711 & 1.816 \\
\rowcolor{gray!10}Ours & \underline{0.702}   & \underline{0.885}  & \underline{0.413}  & \underline{0.764} & \underline{1.096} & \underline{0.761} & \textbf{0.752} & \underline{1.248} & \underline{1.224}  \\
\rowcolor{gray!10}Ours-48 & \textbf{0.759}   & \textbf{0.762}  & \textbf{0.311}  & \textbf{0.824} & \textbf{0.951} & \textbf{0.604}  & \underline{0.748} & \textbf{1.101} & \textbf{1.035}  \\
\bottomrule
\end{tabular}
}
\vspace{-0.2cm}
\caption{\textbf{Quantitative comparisons on LVS6D dataset.}
Here, AP refers to the Average Precision of 3DIoU with a threshold of 50\%. APE denotes the Average Position Error, and ACD represents the Average Chamfer distance.
The results demonstrate that UniPR outperforms Coders across all dataset subsets, with particularly strong improvements observed in the Hard subset with high intra-class variations.
}
\label{tab:all_results}
\vspace{-0.3cm}
\end{table*}

\section{Experiments}
\label{sec:exp}
\vspace{-0.1cm}
\subsection{Experimental Setup}
\noindent
{\bf Dataset} 
In this work, we introduce a \textbf{L}arge-\textbf{V}ocabulary \textbf{S}tereo category-level object dataset for object \textbf{6D} pose estimation and shape reconstruction, named \textbf{LVS6D}. 
This dataset incorporates 3D models from OmniObject3D~\cite{wu2023omniobject3d} and Google Scanned Objects~\cite{downs2022google}. 
We generate approximately 0.4M stereo images for training and 1,000 images for testing, using over 500 HDRI backgrounds. 
To facilitate a more intuitive comparison, we divide the test categories into Easy, Medium, and Hard subsets based on the object complexity.
We compute all evaluation metrics separately for each subset to provide a comprehensive assessment of the proposed method. 
In addition, we capture several real-world scenes using the same stereo camera configuration to evaluate generalization performance, and we conduct real-robot experiments to validate grasping based on the predicted object position and geometry using a simple grasp policy.
Further details are provided in the Appendix.


\noindent
{\bf Implementation Details} 
For the pose-aware shape VAE, we input 2,048 surface points with a feature dimension $\mathbf{C}$ of 256 and a KL divergence channel size $\mathbf{C}_{\rm kl}$ of 64. 
Using objects from the LVS6D dataset, we apply random rotations to all objects, resulting in approximately 76M training pairs. 
UniPR is trained using the AdamW optimizer~\cite{loshchilov2017decoupled} with a weight decay of $10^{-2}$. 
The initial learning rate is set to $2.0 \times 10^{-4}$ and follows a cosine annealing schedule~\cite{loshchilov2016sgdr} for learning rate decay. All experiments are conducted over 24 epochs on 8 RTX 3090 GPUs with a batch size of 8. 
For testing, we use a single RTX 3090 GPU. 



\begin{figure*}[!t]
\centering
\vspace{-0.1cm}
\includegraphics[width=0.85\linewidth]{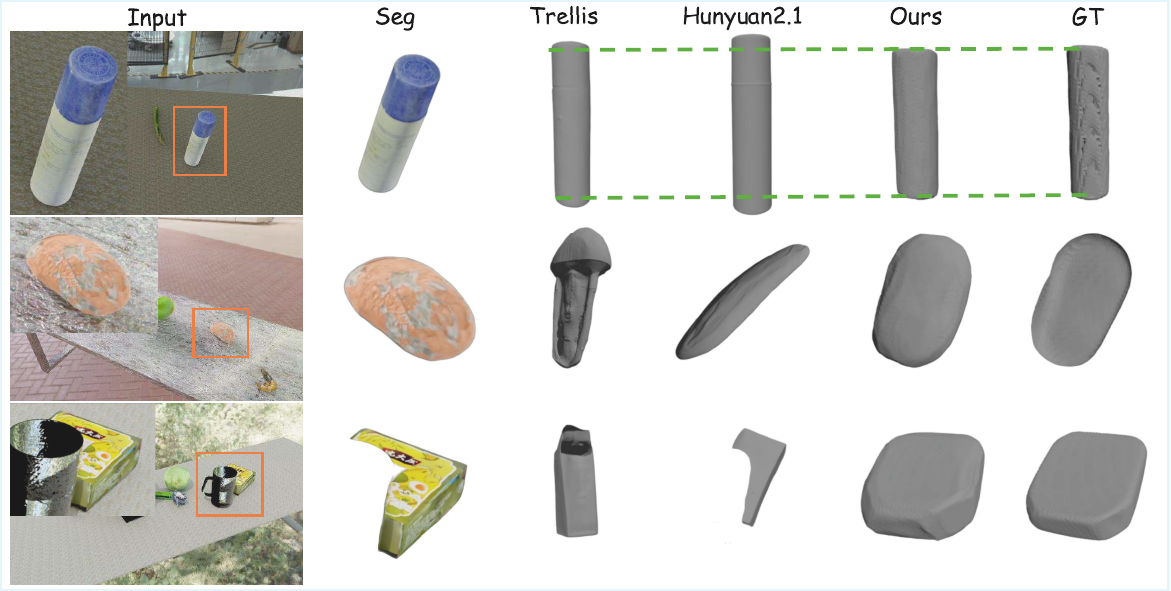}
\vspace{-0.2cm}
\caption{\textbf{Qualitative shape reconstruction results compared with image-to-3D models.} 
The results demonstrate the accurate preservation of shape proportions achieved by our proposed UniPR across various objects in the LVS6D dataset.
}
\vspace{-0.3cm}
\label{fig:vis_gen}
\end{figure*}

\begin{figure*}[!t]
\centering
\includegraphics[width=0.85\linewidth]{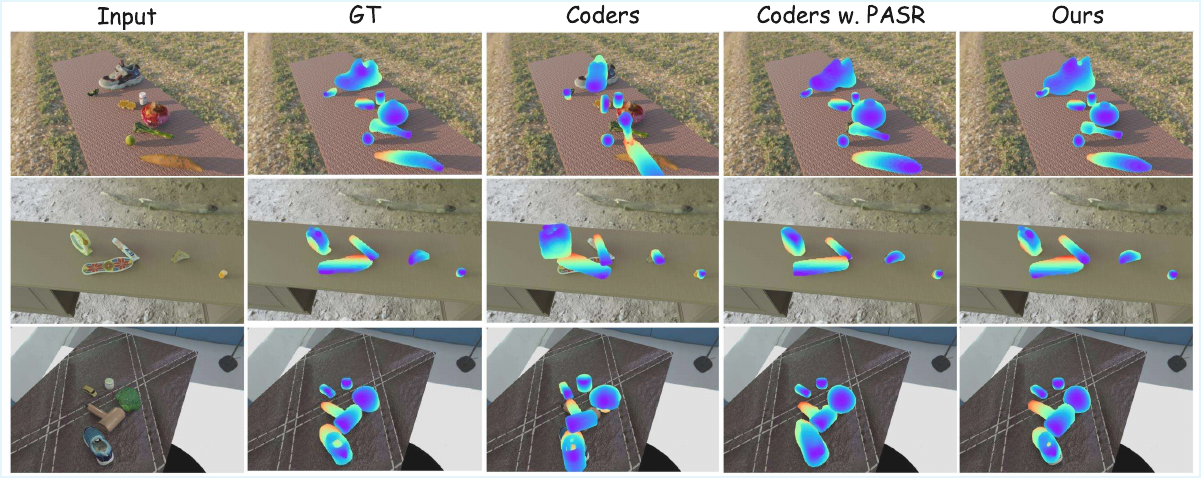}
\vspace{-0.2cm}
\caption{\textbf{Qualitative pose-aware shape reconstruction results on LVS6D dataset.} 
    The results highlight the key role of PASR in simplifying rotation prediction, as it eliminates the ambiguity caused by different canonical definitions for categories with similar geometry.
}
\vspace{-0.3cm}
\label{fig:vis}
\end{figure*}

\vspace{-0.1cm}
\subsection{3D Reconstruction Results}
\vspace{-0.1cm}
\noindent
{\bf Data} 
To evaluate the shape reconstruction performance of our method against SOTA image-to-3D generative models, we select 50 diverse objects from the LVS6D dataset.
For a fair comparison, ground-truth 2D bounding boxes and segmentation masks are provided to all baseline methods.

\noindent
{\bf Metrics} 
We evaluate the quality of reconstructed object meshes using the Chamfer Distance (CD) and F-Score, which measure geometric similarity between the reconstructed and GT meshes.
For all baseline methods, GT poses are provided to ensure accurate shape alignment during evaluation.
In addition, we introduce the Shape Proportion Error (SPE) to assess the physical accuracy of reconstructed objects.
SPE quantifies the relative error in an object’s width, height, and depth, reflecting how well each method preserves true object proportions.

\noindent
{\bf Results} 
The quantitative results are presented in \Cref{tab:gen_result}.
Our method consistently outperforms all baselines across CD, F-Score, and SPE metrics, reflecting the superior geometric fidelity and shape consistency achieved by our approach.
Trellis~\cite{xiang2025structured} and Hunyuan2.1~\cite{hunyuan3d2025hunyuan3d} fails in most cases due to severe occlusions and inaccurate shape proportion estimation.
To provide a fair assessment of its successful cases, we additionally report CD@Top10, which measures the reconstruction quality of its top 10 valid outputs.
Furthermore, we evaluate the efficiency of our end-to-end design, particularly in processing entire scenes containing multiple objects (5 objects in \Cref{tab:gen_result}).
The results demonstrate UniPR achieves up to a \textbf{100×} speed-up in full-scene reconstruction compared to sequential baselines, underscoring its remarkable efficiency in multi-object processing.
We present qualitative comparisons in \Cref{fig:vis_gen}, which highlight the accurate preservation of shape proportions achieved by UniPR and its strong capability for reconstructing previously unseen objects.
Notably, our method performs robustly under occlusion, primarily because it leverages stereo vision and integrates information from the entire image, while others rely only on segmented image regions.

\vspace{-0.1cm}
\subsection{Comparison on LVS6D dataset}
\vspace{-0.1cm}
\noindent
{\bf Metrics} 
We evaluate the performance of the proposed network using three metrics that jointly assess its detection, localization, and reconstruction capabilities.
First, we adopt the Average Precision (AP) based on 3D Intersection over Union (3D IoU) to measure the overall 3D detection performance.
For localization accuracy, we use the Average Position Error (APE), computed as the MSE between the predicted and ground-truth 3D centers.
Finally, to evaluate the quality of pose-aware shape reconstruction, we employ the Average Chamfer Distance (ACD), a widely used metric that measures geometric dissimilarity between reconstructed and ground-truth point sets.

\noindent
{\bf Baselines}
Coders~\cite{zhang2024category} is currently the only stereo-based method designed for category-level 6D object pose estimation and shape reconstruction.
For a fair comparison, we retrain \textit{Coders} on the LVS6D dataset using a canonical shape VAE constructed within a predefined canonical space derived from OmniObject3D.
In addition, we integrate our proposed PASR into Coders as an ablation study to evaluate the effectiveness of our pose-aware design compared to predefined canonical space representations.

\vspace{-0.05cm}
\noindent
{\bf Results} 
\Cref{tab:all_results} presents a comprehensive comparison between our method and existing approaches.
UniPR consistently outperforms all competitors across every evaluation metric and dataset subset, with particularly strong gains observed in the Hard subset.
As object category complexity increases, Coders suffers from greater confusion and degraded performance due to the ambiguities and intra-class variations introduced by predefined canonical spaces.
In contrast, the integration of our PASR into Coders further validates the advantages of our design as it effectively mitigates the ambiguities inherent in human-defined canonical spaces and facilitates scalable training on large-vocabulary datasets.
Qualitative comparisons shown in \Cref{fig:vis} further support these findings.
Notably, the rotation predictions of UniPR exhibit significant improvement over Coders, primarily because our pose-aware design eliminates rotational ambiguities for categories with similar geometric structures.

\vspace{-0.1cm}
\subsection{Comparison on Public Stereo datasets}
\vspace{-0.1cm}
\noindent
{\bf Data} 
To further investigate the our design for stereo vision, we further introduce the public TOD~\cite{liu2020keypose} and SS3D~\cite{zhang2024category} datasets for evaluation.
We follow the same settings with Coders~\cite{zhang2024category} and train UniPR separately on each of the two datasets and evaluate our method on novel instances from each category that are not included in the training process.

\noindent
{\bf Metrics} 
Following~\cite{zhang2024category}, we adopt the 3D IoU, which jointly evaluates the accuracy of rotation, translation and scale.
In addition, we assess rotation and translation precision separately using threshold-based accuracy metrics.
Specifically, a prediction is regarded as correct only if rotation error falls within $\{5^{\circ}, 10^{\circ}\}$ and translation error lies within $\{2 cm,5 cm,10 cm\}$.
This combined evaluation provides a comprehensive measure of spatial accuracy.

 \noindent
{\bf Pose estimation} 
For a fair comparison with previous stereo-based pose estimation methods, we introduce an additional pose decoding layer following the pose embeddings.
Given a canonical shape definition during training, UniPR predicts the relative object pose using several MLP layers.

 \noindent
{\bf Results} 
Our proposed method outperforms most others across various metrics, demonstrating the significant superiority of our proposed modules, as shown in \cref{tab:compari_public}. 
The TOD dataset emphasizes transparent objects, highlighting the importance of stereo vision due to the absence of depth sensor data. 
Our method also surpasses competitors on the SS3D dataset, showcasing the efficiency of our pose-aware shape representation. 
Additional detailed and quantitative results are provided in the supplementary material.

\vspace{-0.1cm}
\subsection{Ablation Study}
\vspace{-0.1cm}
\noindent
We present ablation experiments on the LVS6D dataset.

\noindent
{\bf Insight of Pose-aware Shape Representation} 
In this experiment, we aim to explicitly validate the effectiveness of the proposed PASR.
To this end, we perform a comparative analysis by replacing PASR with a conventional canonical shape reconstruction while keeping the rest of the UniPR architecture unchanged.
The results, presented in \Cref{tab:all_ablation_results}, clearly demonstrate that PASR is critical for achieving high reconstruction accuracy.
When PASR is removed, the ACD on challenging subsets increases by nearly \textbf{10x}, indicating that previous canonical representations fail to establish robust category priors for geometrically diverse objects.

\noindent
{\bf Stereo or Monocular}
We conduct experiments to examine the advantages of using stereo cameras over monocular ones.
For this comparison, we train and evaluate a variant of our model using only the left-view images from the stereo setup.
As shown in the first columns of \Cref{tab:all_ablation_results}, the stereo configuration significantly outperforms the monocular counterpart.
This improvement primarily stems from the inherent limitation of monocular inputs in recovering accurate depth information, particularly for objects with diverse scales or complex spatial arrangements.

\noindent
{\bf Impact of Spherical Voxel Space}
As discussed in \Cref{sec:method_pasa}, we utilize a spherical voxel space instead of a cubic voxel space to maintain rotational consistency.
This design is essential because normalizing objects with varying rotations can introduce unwanted scale dependencies linked to their orientation, leading to confusion during training. 
To evaluate the effectiveness of this approach, we conducted an experiment comparing spherical and cubic voxel spaces on the hard subset of LVS6D, where complex objects frequently encounter these issues.
\Cref{tab:ablation_spherical} demonstrates that our proposed spherical voxel space significantly enhances training stability and performance.

\begin{table}[!t]
\centering
\resizebox{\linewidth}{!}{
\begin{tabular}{cl|llllll}
\toprule
\textbf{Dataset}              & \textbf{Method}      & $3D_{25}$ & $3D_{50}$ & $3D_{75}$ & $\makecell[c]{10^{\circ}\\10\mathrm{cm}}$ & $\makecell[c]{10^{\circ}\\5\mathrm{cm}}$  & $\makecell[c]{5^{\circ}\\2\mathrm{cm}}$   \\ 
\midrule
                      & SPD~\cite{tian2020shape}         & 54.1 & 13.6 & -    & 11  & 6.9    & -     \\
                      & SGPASPD~\cite{chen2021sgpa}       & 55.8 & 14.6 & -    & 13.8  & 8.1    & -     \\
                      & KeyPose~\cite{liu2020keypose}    & -    & -    & -    & 43.7 & 38.7    & -     \\
                      & StereoPose~\cite{chen2023stereopose} & 91.7 & 49.8 & -    & 54.2 & 46.1    & -     \\
                      & Coders~\cite{zhang2024category}      & 100  & \textbf{99.8} & \textbf{62.1} & \textbf{90.8} & \textbf{90.8}    & \textbf{64.8}  \\
\rowcolor{gray!10} \multirow{-6}{*}{\centering {TOD~\cite{liu2020keypose}}} & Ours       & \textbf{100}  & 97.5     &52.5      &86.9      &86.9     &63.2       \\ 
\midrule
\textbf{Dataset}              & \textbf{Method}      & $3D_{25}$ & $3D_{50}$ & $3D_{75}$ & $\makecell[c]{10^{\circ}\\10\mathrm{cm}}$ & $\makecell[c]{5^{\circ}\\5\mathrm{cm}}$  & $\makecell[c]{5^{\circ}\\2\mathrm{cm}}$   \\ 
\midrule
                      & Coders~\cite{zhang2024category}      & -    & 61.7 & 21.5 & -    & 54.7 & 28.1  \\
\rowcolor{gray!10} \multirow{-2}{*}{\centering {SS3D~\cite{zhang2024category}}} & Ours       & \textbf{91.5} & \textbf{89.3} & \textbf{55.4} & \textbf{88.4} & \textbf{78.4} & \textbf{41.3}  \\
\bottomrule
\end{tabular}
}
\vspace{-0.25cm}
\caption{\textbf{Comparison on public stereo datasets.}
The results highlight the effectiveness of the proposed PASR and its capability to decode accurate 6D object poses.
}
\label{tab:compari_public}
\vspace{-0.3cm}
\end{table}

\begin{table}[!t]
\centering
\resizebox{\linewidth}{!}{
\begin{tabular}{l|cc|cc|cc}
\toprule
\multirow{2}{*}{\textbf{Method} }&\multicolumn{2}{c|}{\textbf{Easy}} &\multicolumn{2}{c|}{\textbf{Medium}}   &\multicolumn{2}{c}{\textbf{Hard}}  \\
     & AP$\uparrow$ & ACD$\downarrow$ & AP$\uparrow$ & ACD$\downarrow$ & AP$\uparrow$ & ACD$\downarrow$     \\
\midrule
w.o. Pose-aware S. R. & 0.370   & 1.258  & 0.330  & 2.711 & 0.196  & 12.363  \\
\rowcolor{gray!10}Ours & \textbf{0.702}    & \textbf{0.413}  & \textbf{0.764}  & \textbf{0.761} & \textbf{0.752} & \textbf{1.224} \\
\midrule
Monocular & 0.173    & 1.676 & 0.266   & 1.633  & 0.270   & 2.444   \\
\rowcolor{gray!10}Stereo & \textbf{0.702}    & \textbf{0.413}  & \textbf{0.764}  & \textbf{0.761} & \textbf{0.752} & \textbf{1.224} \\
\bottomrule
\end{tabular}
}
\vspace{-0.25cm}
\caption{\textbf{Ablation study results.}
The results underscore the critical role of PASR in enabling large-vocabulary object detection and pose-aware shape reconstruction, as well as the importance of the stereo design for achieving accurate geometric recovery.
}
\label{tab:all_ablation_results}
\vspace{-0.4cm}
\end{table}


\begin{table}[!t]
\centering
\vspace{-0.1cm}
\resizebox{0.75\linewidth}{!}{
\begin{tabular}{l|ccc}
\toprule
\multirow{2}{*}{\textbf{Method} }&\multicolumn{3}{c}{\textbf{Hard}}  \\
     & AP$\uparrow$ & APE$\downarrow$ & ACD$\downarrow$     \\
\midrule
w.o. Spherical Voxel Space & 0.677          & \textbf{1.129}          & 1.310 \\
\rowcolor{gray!10}Ours                       & \textbf{0.752} & 1.248 & \textbf{1.224}\\
\bottomrule
\end{tabular}
}
\vspace{-0.2cm}
\caption{\textbf{Ablation on spherical voxel space.}
We conduct this experiment with shape decoder utilizing spherical and cubic voxel space. This experiment is conducted only on the hard subset of LVS6D, as most re-normalized objects are included in this subset. 
}
\label{tab:ablation_spherical}
\vspace{-0.5cm}
\end{table}

%% file: sec/5_con.tex
\vspace{-0.2cm}
\section{Conclusion}
\label{sec:con}
\vspace{-0.2cm}

In this paper, we propose the first unified object-level real-to-sim transfer network, UniPR, designed for improved perception and reconstruction from a single stereo image pair.
We construct a large-vocabulary stereo dataset, LVS6D, comprising 192 categories and over 6,300 everyday objects, leveraging the efficiency of our Pose-Aware Shape Representation (PASR).
Furthermore, we establish a unified paradigm that eliminates the need for separate detection, segmentation, and reconstruction submodules, enabling whole-scene object processing in a single feedforward pass.
Experiments demonstrate that UniPR achieves accurate shape proportions and accelerated inference speed, benefiting from the integrated PASR design.
The principal limitation of UniPR is its reduced performance in monocular settings due to insufficient depth information.
We aim to address this limitation by incorporating large-scale depth modules, which would allow our proposed representation to function effectively with simpler camera setups.


%% file: sec/X_suppl.tex
\clearpage
\setcounter{page}{1}
\maketitlesupplementary

\section{Overview}

We provide \textbf{videos} that offer a brief introduction to UniPR, along with additional visualizations on real-world data and real-robot experiments. All video files are included in the Supplementary Material.

The formulation and implementation details of the Pose-Aware Shape VAE are provided in \cref{sec:sup-vae}.
Similarly, further details of the LVS6D dataset are described in \cref{sec:sup-lvs6d}.
More ablation studies are provided in \cref{sec:add_exp}.
With the emergence of SAM 3D Objects, we also clarify the conceptual and technical differences between their approach and UniPR, and present several comparative examples in \cref{sec:sup-sam}.
Real-world experimental results of UniPR are presented in \cref{sec:sup-real}.
For a more comprehensive understanding of reconstruction quality, we include visual comparisons between Coders~\cite{zhang2024category} and UniPR in \cref{sec:sup-recon}.
Moreover, qualitative results on the proposed LVS6D dataset are summarized in \cref{sec:vis-lvs6d}.
Finally, detailed evaluation metrics and visualizations on public datasets are provided in \cref{sec:sup-public}.

\begin{figure*}[htb]
    \centering
    \includegraphics[width=0.85\linewidth]{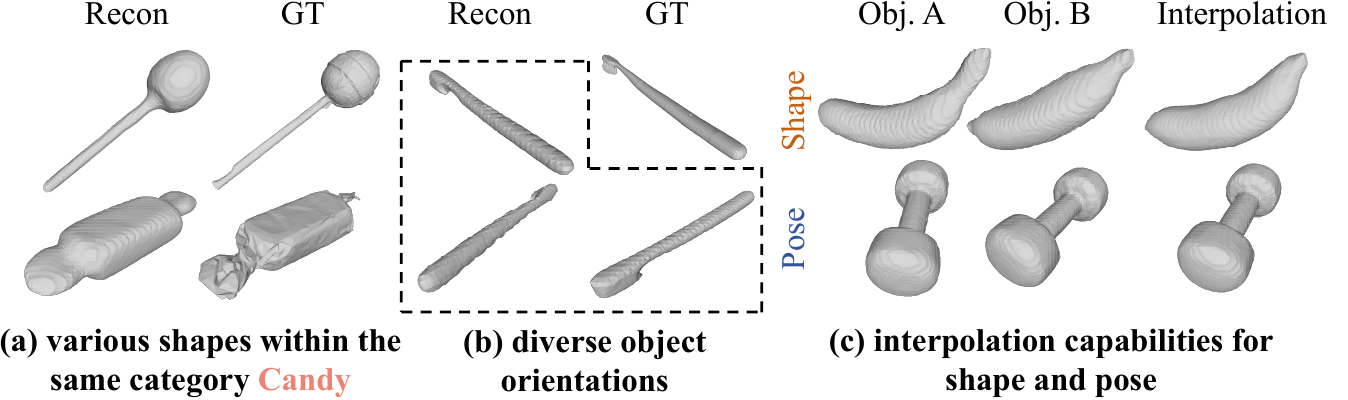}
    \caption{\textbf{Generative performance of PASR.} We evaluate PASR’s generative performance on same-category shapes (a), diverse object orientations (b), and pose-shape interpolation capabilities (c). Results demonstrate the robustness of the learned embedding space.}
    \label{fig:gen_pasr}
\end{figure*}

\vspace{-0.2cm}
\section{Details for PASR Design}\label{sec:sup-vae}
\vspace{-0.1cm}
\noindent \textbf{Details for Pose-Aware Shape VAE}
Our VAE model is based on 3DShape2VecSet~\cite{zhang20233dshape2vecset} and takes object surface point clouds in an object-centered coordinate system as input. We adopt a spherical voxel space by sampling query points from a unit sphere rather than a cube, which is more suitable for representing objects under diverse rotations. 
The objective of our VAE model is to encode objects into lightweight embeddings for use in the detection pipeline. All ground-truth shape distributions for detection pipeline are generated using the encoder of our pretrained VAE model.

For the encoder $\varepsilon_{\rm VAE}$ we first utilize self-attention layers to extract information from $\boldsymbol{z}_{\rm surface} \in \mathbb{R}^{N \times C}$ to point latent embedding $\boldsymbol{z}_{\rm point} \in \mathbb{R}^{N \times C}$. 
\begin{equation}
\boldsymbol{z}_{\rm point} = {\rm SelfAttn}(\boldsymbol{z}_{\rm surface})
\end{equation}
We initialize the object latent embedding, $\boldsymbol{z}_{\rm object} \in \mathbb{R}^{C}$, and employ cross-attention layers to learn a mapping from the point distribution to the object embedding, as described in the main text. To ensure an effective representation and reduce the latent channel, we apply KL regularization during this process.

For the decoder $\mathcal{D}_{\rm VAE}$, we first generate the sampled object latent embedding $\boldsymbol{z}_{\rm sampled} \in \mathbb{R}^{C}$, as outlined in the main text. Next, we initialize the point latent embedding, $\boldsymbol{\hat{z}}_{\rm point} \in \mathbb{R}^{N \times C}$, using a standard Gaussian distribution. Cross-attention layers are then utilized to recover information from the sampled object latent embedding $\boldsymbol{z}_{\rm sampled}$ into the point latent embedding:
\begin{equation}
\boldsymbol{\hat{z}}_{\rm point} = {\rm CrossAttn}(\boldsymbol{\hat{z}}_{\rm point}, \boldsymbol{z}_{\rm sampled})
\end{equation}
Further architectural details of the proposed pose-aware shape VAE are provided in the main text.


\noindent \textbf{Generative performance of PASR.}
We evaluate the generative performance of PASR across various shapes within the same category, diverse object orientations, and its interpolation capabilities in both pose and shape, as illustrated in \cref{fig:gen_pasr}. The results demonstrate the robustness of the learned PASR embedding space. 

\noindent \textbf{Texture Retrieval.}
Our method focuses primarily on shape reconstruction and does not include a dedicated texture generation module. However, given the VecSet representation produced by UniPR, we can seamlessly integrate external texture generation models. In particular, we employ Hunyuan3D-Paint-v2.1 to synthesize high-quality textures based on our reconstructed shapes, as demonstrated in the teaser, \cref{fig:vis_sam_real} and \cref{fig:vis_sam_sim}.

\begin{figure*}[tb]
  \centering
  \includegraphics[width=1\linewidth]{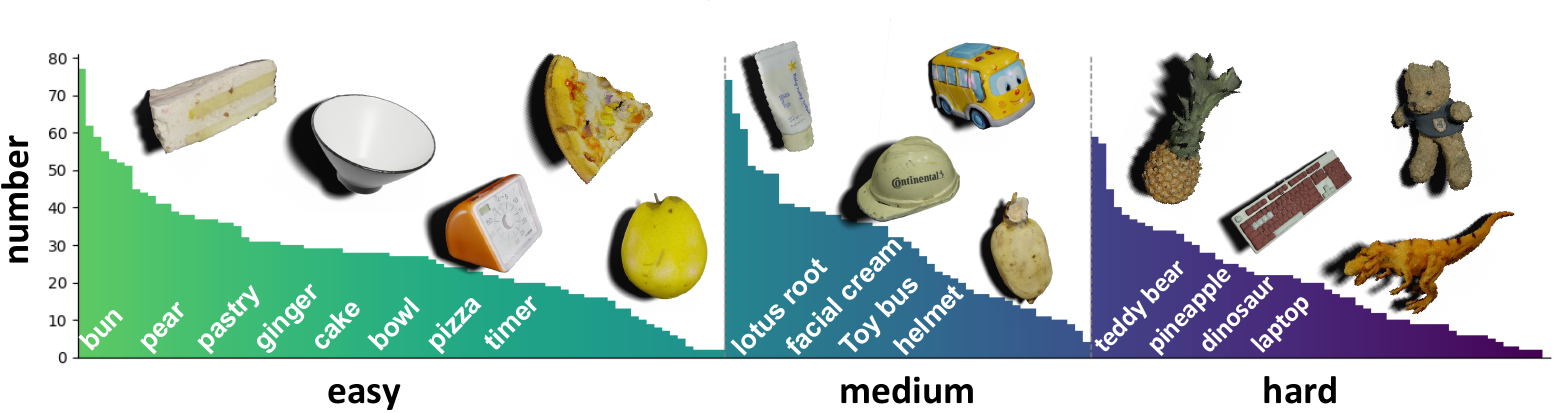}
  \caption{
  \textbf{Category Distribution of the LVS6D dataset.}
    We divide the dataset into three subsets Easy, Medium, and Hard based on the reconstruction difficulty. As the difficulty increases, the objects in each subset exhibit more complex geometries.
  }
  \label{fig:dataset}
\end{figure*}

\vspace{-0.2cm}
\section{Details for LVS6D}
\label{sec:sup-lvs6d}
\vspace{-0.1cm}
\noindent
{\bf Dataset structure.} 
Our dataset is organized using a folder-based structure.
For each stereo image, we provide object masks along with comprehensive annotations, including object category, position, scale, rotation, and the corresponding 3D shape. The object shape data include textured meshes sourced from OmniObject3D~\cite{wu2023omniobject3d} and Google Scanned Objects (GSO)~\cite{downs2022google}. OmniObject3D contains 6,000 scanned objects across 190 daily-use categories, while GSO provides over 1,000 high-quality 3D-scanned household items.

The proposed LVS6D dataset spans 192 categories and includes more than 6,300 scanned objects. We generate approximately 0.4M stereo images for training and 1,000 images for testing, using over 500 high–dynamic-range (HDRI) backgrounds. Rendering is performed with BlenderProc 2.6.1~\cite{denninger2023blenderproc2}. The stereo camera configuration uses a 13,cm baseline, and all images are captured at a resolution of 1920$\times$1200 pixels.

In addition to synthetic data, we also capture several real-scene images using the identical stereo setup to evaluate real-world generalization.
Overall, LVS6D is, to our knowledge, the largest stereo category-level object dataset designed for 6D pose estimation and shape reconstruction to date.

\begin{table*}[h]
\centering
\resizebox{0.95\linewidth}{!}{
\begin{tabular}{l|l}
\toprule
\textbf{Subsets} & Object Samples \\
\midrule
\textbf{Easy} & peach, ball, cake, bread, bun, egg, medicine bottle, mango, pomegranate, walnut \\
\textbf{Medium} & biscuit, candy, cheese, chili, doll, lotus root, small box, shoe, soap, teapot \\
\textbf{Hard} & pineapple, dinosaur, razor, teddy bear, keyboard, laptop, bamboo shoots, banana, clock, glasses case \\
\bottomrule
\end{tabular}
}
\caption{\textbf{Representative categories of LVS6D across subsets.} Ten representative categories are listed per subset: Easy subset objects have simple regular geometries, while Medium/Hard subsets include increasingly complex shapes and higher intra-class variability.}
\label{tab:subsets}
\end{table*}

\noindent
{\bf LVS6D subsets.} 
As discussed in the main text, we divide the categories in LVS6D into Easy, Medium, and Hard subsets according to object complexity. \cref{tab:subsets} lists ten representative categories for each subset. Objects in the Easy subset exhibit simple and regular geometries (e.g., spheres, cubes) and display relatively low intra-class variation. In contrast, the Medium and Hard subsets contain objects with increasingly complex shapes and significantly higher intra-class variability, as shown in \cref{fig:dataset}.
For all evaluation metrics, we compute the results for every category individually and report the mean values for each subset.

\section{Additional Experimental Results}\label{sec:add_exp}

\subsection{Ablation Study on Loss Function}
The loss term $\mathcal{L}_{\text{shape}}$ comprises both the KL divergence loss $\mathcal{L}_{\text{kl}}$ and the reconstruction loss $\mathcal{L}_{\text{recon}}$, as defined in Eq.7. We also conducted an ablation study without KL-based supervision, as shown in \cref{tab:kl}. The results demonstrate the importance of KL-based supervision for utilizing the pretrained VAE model.

\begin{table}[htb]
    \centering
    \resizebox{0.95\linewidth}{!}{
    \begin{tabular}{l|ccc}
    \toprule
    \textbf{Method} & AP$\uparrow$ & APE$\downarrow$ & ACD$\downarrow$     \\
    \midrule
    w.o. KL-based supervision & 0.675          & 1.210         & 2.947 \\
    \rowcolor{gray!10}Ours    & \textbf{0.752} & \textbf{1.248} & \textbf{1.224}\\
    \bottomrule
    \end{tabular}
    }
    \caption{\textbf{The ablation of KL-based supervision.} The results demonstrate the importance of KL-based supervision for utilizing the pretrained VAE model.}
    \label{tab:kl}
\end{table}

\subsection{Fairness Comparison on Input Modalities}
When comparing stereo methods with monocular baselines like Trellis, a potential concern is the inherent scale ambiguity present in monocular inputs. 
To verify that UniPR's performance gains stem primarily from architectural innovation rather than merely the sensor modality, we conducted extended experiments across four configurations: monocular Trellis, Trellis-stereo (extending Trellis with stereo input using ground-truth crops), Ours-mono (our pipeline adapted for monocular input), and our full stereo UniPR. 

As shown in \cref{tab:fairness_result}, UniPR-mono significantly outperforms both Trellis variants in geometric consistency (CD and F-Score). Furthermore, our full stereo model exceeds Trellis-stereo by a wide margin in Shape Proportion Error (SPE). This confirms that our proposed Pose-Aware Shape Representation and the end-to-end Triplane integration are the primary drivers of the observed improvements, demonstrating that UniPR efficiently fuses geometric constraints to achieve superior metric-scale reconstruction.

\begin{table}[htb]
\centering
\resizebox{0.95\linewidth}{!}{
\begin{tabular}{l|c|c|c|c}
\toprule
\textbf{Method} & \textbf{Input Modality} & \textbf{CD $\downarrow$} & \textbf{F-Score $\uparrow$} & \textbf{SPE $\downarrow$} \\
\midrule
Trellis  & Monocular & 0.1096 & 0.334 & 0.475 \\
Trellis-stereo & Stereo (GT Crops) & 0.1001 & 0.388 & 0.376 \\
\midrule
Ours-mono & Monocular & \underline{0.0141} & \underline{0.868} & \underline{0.110} \\
\rowcolor{gray!10} Ours (Full)  & \textbf{Stereo} & \textbf{0.0083} & \textbf{0.883} & \textbf{0.109} \\
\bottomrule
\end{tabular}
}
\caption{
\textbf{Fairness comparison on input modalities and reconstruction metrics.} 
Our method demonstrates superior geometric consistency and shape proportion accuracy compared to Trellis, regardless of the input modality.
}
\label{tab:fairness_result}
\end{table}

\subsection{Part-Aware Refinement on TOD}

In evaluating our method on the TOD dataset, we observed a performance discrepancy in categories with complex topologies, such as the "mug" category. The primary cause of this gap is that baseline methods like Coders~\cite{zhang2024category} employ a part-aware prediction strategy to determine rotation based on specific features (e.g., handle positioning). In contrast, our original pipeline relied on global orientation without explicit part-level priors. Furthermore, another primary driver of the observed performance gap is that the baseline models are pre-trained on the comprehensive SS3D dataset, whereas our model is trained from scratch.

To ensure a rigorous and fair comparison, we implemented a part-aware refinement stage specifically for mug-like objects. As shown in \cref{tab:tod_refinement}, this enhancement enables UniPR to effectively resolve topological complexities and outperform Coders across most overall evaluation metrics.

\begin{table}[htb]
\centering
\resizebox{0.95\linewidth}{!}{
\begin{tabular}{l|c|cc|cc}
\toprule
\multirow{2}{*}{\textbf{Method}} & \multirow{2}{*}{\textbf{Refinement}} & \multicolumn{2}{c|}{\textbf{Mug}} & \multicolumn{2}{c}{\textbf{Overall}} \\
& & 5$^\circ$ 2cm $\uparrow$ & 10$^\circ$ 5cm $\uparrow$ & 5$^\circ$ 2cm $\uparrow$ & 10$^\circ$ 5cm $\uparrow$ \\
\midrule
Coders & Part-Aware & \textbf{56.2} & 81.9 & 64.8 & 90.8 \\
\midrule
Ours (Original) & Global-only & 36.8 & 74.6 & 63.2 & 86.9 \\
\rowcolor{gray!10} Ours (\textbf{Enhanced}) & \textbf{Part-Aware} & 46.6 & \textbf{95.0} & \textbf{69.1} & \textbf{97.5} \\
\bottomrule
\end{tabular}
}
\caption{
\textbf{Evaluation on the TOD dataset with part refinement.} 
By incorporating part-aware refinement, our enhanced model successfully resolves topological complexities and maintains its superiority in real-to-sim perception tasks.
}
\label{tab:tod_refinement}
\end{table}

\section{Comparison with SAM 3D Objects}
\label{sec:sup-sam}

SAM 3D Objects is a recent image-to-3D object generation pipeline developed concurrently with UniPR. Here, we clarify the key differences.

First, although SAM 3D unifies segmentation and reconstruction, it still relies on an external detection module for object perception. In contrast, UniPR provides a fully end-to-end perception–reconstruction pipeline without requiring separate detection or segmentation stages.

\begin{figure*}[htb]
    \centering
    \includegraphics[width=0.9\linewidth]{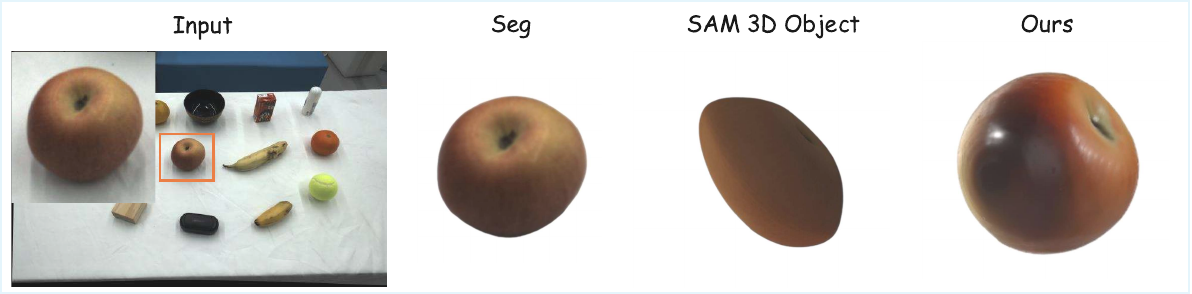}
    \caption{Comparision with SAM 3D Object on real scene.}
    \label{fig:vis_sam_real}
\end{figure*}

\begin{figure*}[htb]
    \centering
    \includegraphics[width=0.9\linewidth]{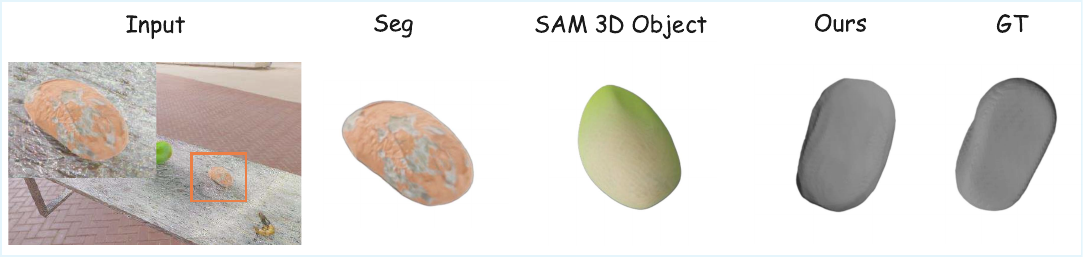}
    \caption{Comparision with SAM 3D Object on simulated scene.}
    \label{fig:vis_sam_sim}
\end{figure*}

\begin{figure}[htb]
    \centering
    \includegraphics[width=1\linewidth]{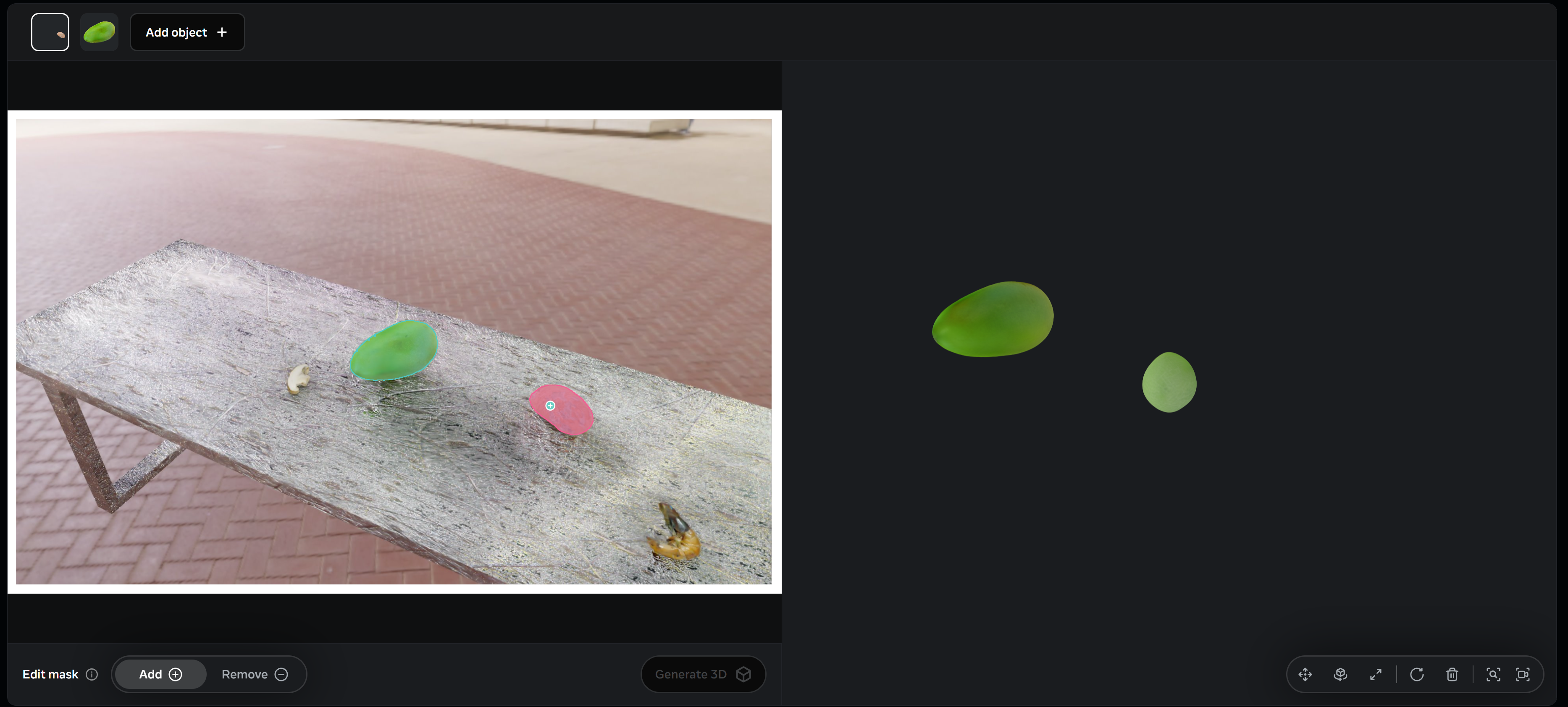}
    \caption{Illustration of SAM 3D Object test result.}
    \label{fig:vis_sam_test}
\end{figure}

Second, UniPR leverages stereo vision, whereas SAM 3D operates on monocular images. Stereo geometry provides metric depth cues, enabling accurate real-to-sim transfer and supporting real-world robotic grasping experiments, as demonstrated in \cref{sec:sup-real}. This depth-aware design also leads to more reliable shape proportions in many cases and allows UniPR to handle objects with ambiguous or hard-to-recognize categories (e.g., stones and unstructured shapes) as shown in \cref{fig:vis_sam_real}, \cref{fig:vis_sam_sim} and \cref{fig:vis_sam_test}.

Third, SAM 3D processes objects individually, requiring one-by-one reconstruction. UniPR, by contrast, performs parallel full-scene processing in a single forward pass, offering significantly higher efficiency and better utilization of global context.

In summary, UniPR differs fundamentally from SAM 3D Objects. Rather than serving as a general-purpose image-to-3D generator, UniPR targets tabletop object perception, metric reconstruction, and downstream robotic manipulation, making it a specialized and practically oriented model within the real-to-sim domain.

\section{Results on Real-world Data}
\label{sec:sup-real}
\begin{figure*}[!t]
\setlength\tabcolsep{0.5 pt}
\centering
\scalebox{0.9}{
\begin{tabular}{cccc}

Input-Left & Input-Right & Coders~\cite{zhang2024category} & UniPR \\

\begin{tabular}{l}\includegraphics[width=0.25\linewidth]{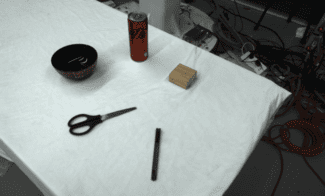}\end{tabular} &
\begin{tabular}{l}\includegraphics[width=0.25\linewidth]{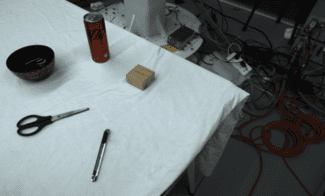}\end{tabular} &
\begin{tabular}{l}\includegraphics[width=0.25\linewidth]{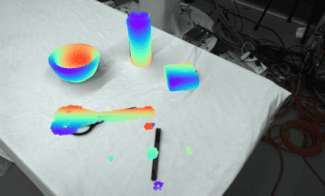}\end{tabular} &
\begin{tabular}{l}\includegraphics[width=0.25\linewidth]{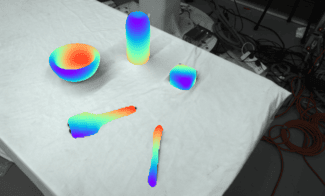}\end{tabular} \\

\begin{tabular}{l}\includegraphics[width=0.25\linewidth]{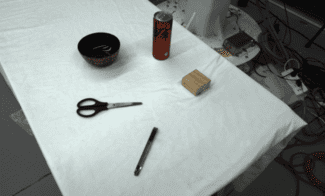}\end{tabular} &
\begin{tabular}{l}\includegraphics[width=0.25\linewidth]{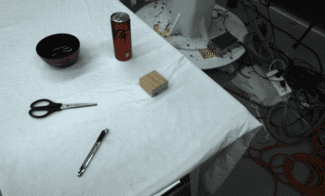}\end{tabular} &
\begin{tabular}{l}\includegraphics[width=0.25\linewidth]{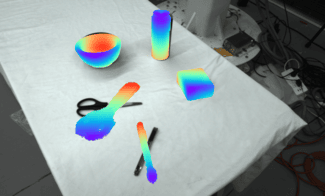}\end{tabular} &
\begin{tabular}{l}\includegraphics[width=0.25\linewidth]{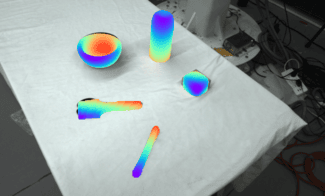}\end{tabular} \\

\begin{tabular}{l}\includegraphics[width=0.25\linewidth]{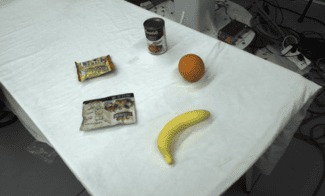}\end{tabular} &
\begin{tabular}{l}\includegraphics[width=0.25\linewidth]{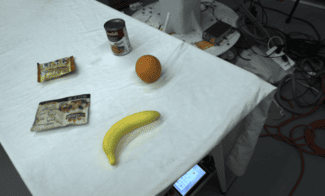}\end{tabular} &
\begin{tabular}{l}\includegraphics[width=0.25\linewidth]{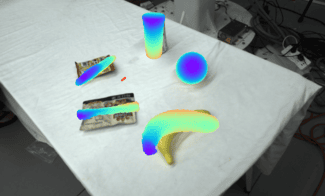}\end{tabular} &
\begin{tabular}{l}\includegraphics[width=0.25\linewidth]{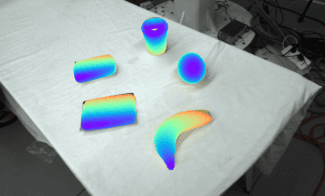}\end{tabular} \\

\begin{tabular}{l}\includegraphics[width=0.25\linewidth]{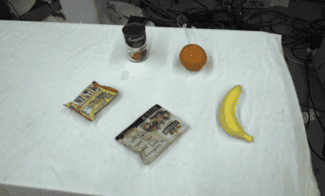}\end{tabular} &
\begin{tabular}{l}\includegraphics[width=0.25\linewidth]{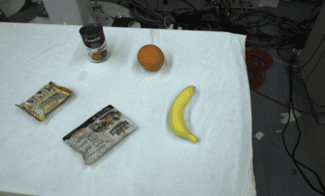}\end{tabular} &
\begin{tabular}{l}\includegraphics[width=0.25\linewidth]{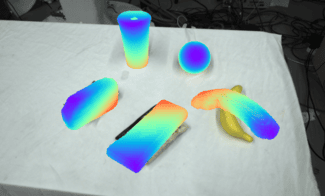}\end{tabular} &
\begin{tabular}{l}\includegraphics[width=0.25\linewidth]{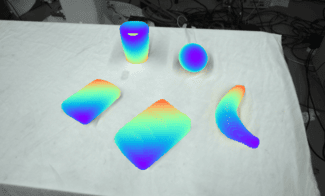}\end{tabular} \\

\end{tabular}}
\caption{\textbf{Qualitative results on real-world data.} 
    We present visualizations of Coders and UniPR on real-world data. We compare the UniPR with Coders~\cite{zhang2024category}. The results demonstrate UniPR's strong generalization ability with real-world data.
}
\label{fig:lvs_results_real}
\end{figure*}

We further present qualitative comparisons on real-world scenes in \cref{fig:lvs_results_real}, where we compare UniPR with Coders. The results demonstrate that UniPR exhibits strong generalization capability when applied to real-world inputs.

To further validate the metric accuracy of UniPR’s predicted object positions and scales, we conduct real-robot experiments using a simple top-to-bottom grasping policy. 
By directly using UniPR’s metric-scale pose and shape predictions as input to the grasping policy, the robot successfully grasps a variety of objects, confirming the practical reliability of our method. Grasping results are provided in the accompanying video.

\section{Visualization of Reconstruction Results}
\label{sec:sup-recon}
\begin{figure*}[!t]
\setlength\tabcolsep{0.5 pt}
\centering
\includegraphics[width=0.6\linewidth]{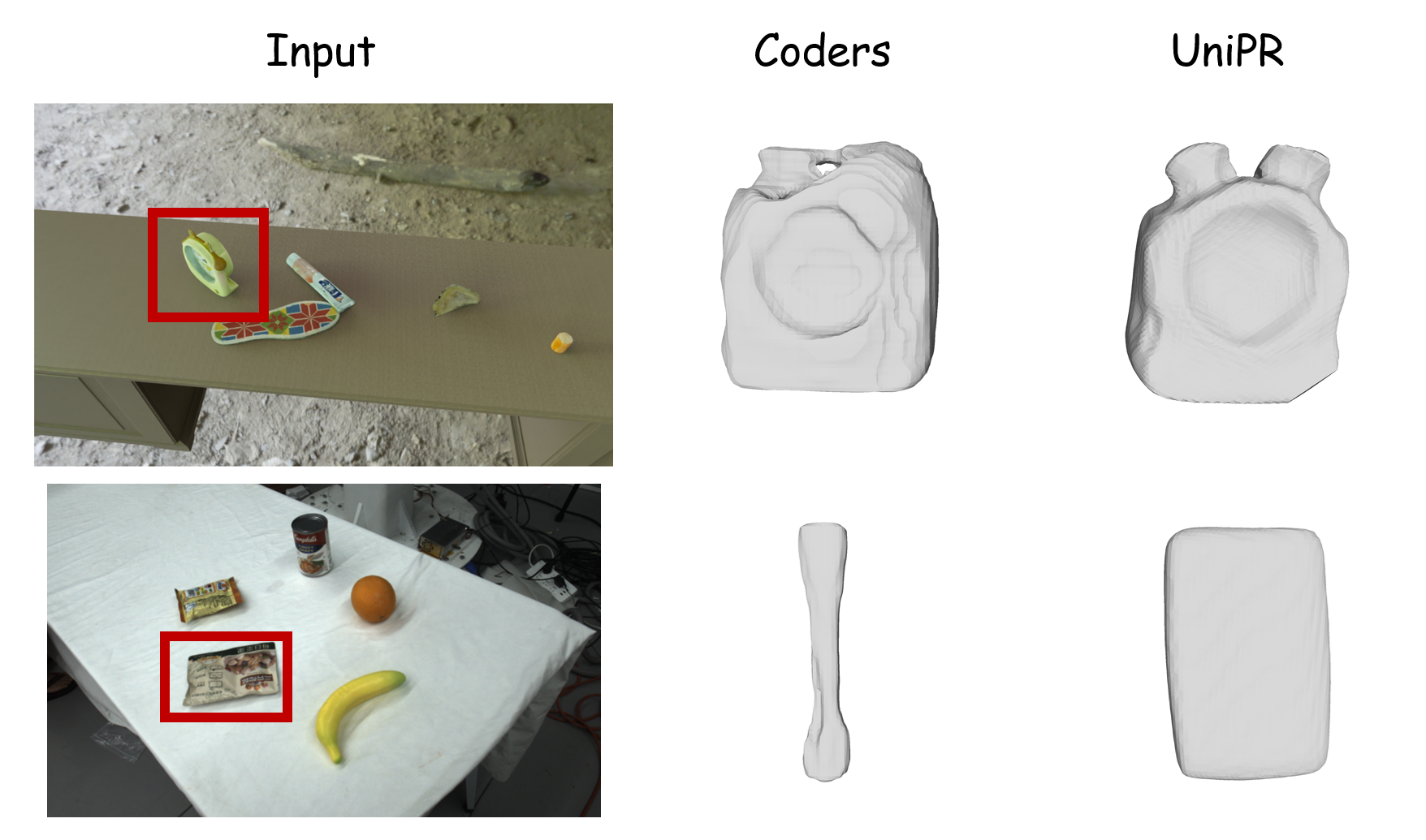}
\caption{
\textbf{Qualitative Results of Reconstruction.} Our approach generates meshes with high quality which is attributed to the effectiveness of pose-aware shape representation.
}
\label{fig:lvs_recon}
\end{figure*}

We present the visualization of reconstruction results on the LVS6D dataset in \cref{fig:lvs_recon}. As demonstrated, our UniPR method consistently produces high-quality meshes, showcasing the effectiveness of pose-aware shape reconstruction. Our UniPR does not need category-level prior so we can generate a more detailed object mesh.

\section{Qualitative Results on LVS6D}
\label{sec:vis-lvs6d}
\begin{figure*}[!t]
\setlength\tabcolsep{0.5 pt}
\centering
\scalebox{0.9}{
\begin{tabular}{cccc}

Input & GT & Coders~\cite{zhang2024category} & UniPR \\

\begin{tabular}{l}\includegraphics[width=0.25\linewidth]{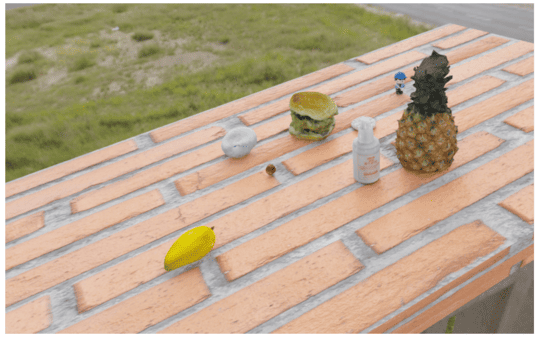}\end{tabular} &
\begin{tabular}{l}\includegraphics[width=0.25\linewidth]{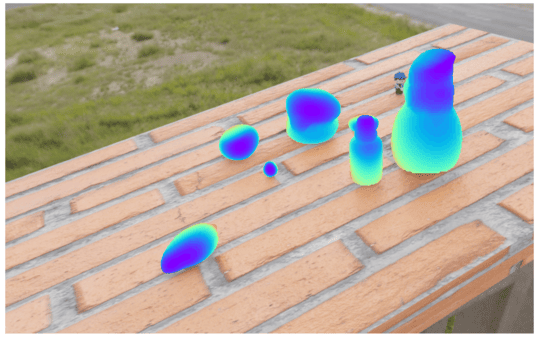}\end{tabular} &
\begin{tabular}{l}\includegraphics[width=0.25\linewidth]{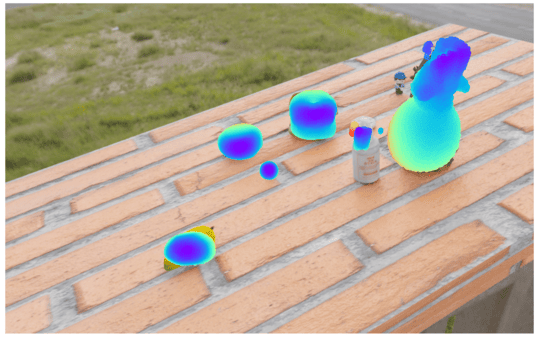}\end{tabular} &
\begin{tabular}{l}\includegraphics[width=0.25\linewidth]{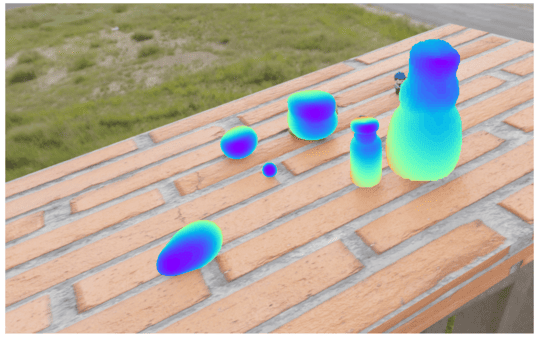}\end{tabular} \\

\begin{tabular}{l}\includegraphics[width=0.25\linewidth]{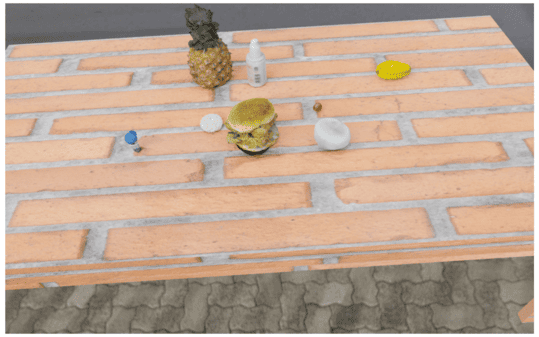}\end{tabular} &
\begin{tabular}{l}\includegraphics[width=0.25\linewidth]{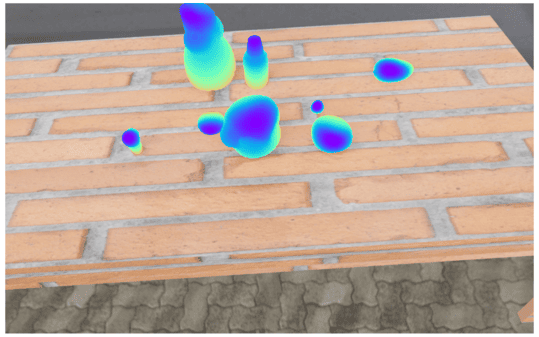}\end{tabular} &
\begin{tabular}{l}\includegraphics[width=0.25\linewidth]{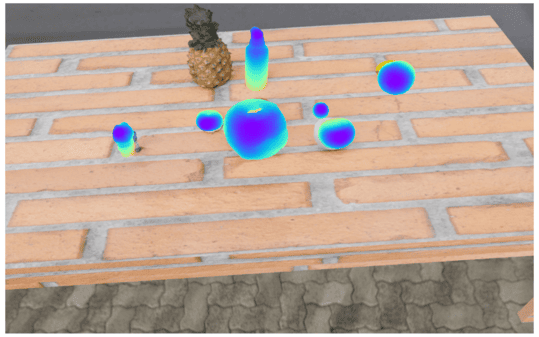}\end{tabular} &
\begin{tabular}{l}\includegraphics[width=0.25\linewidth]{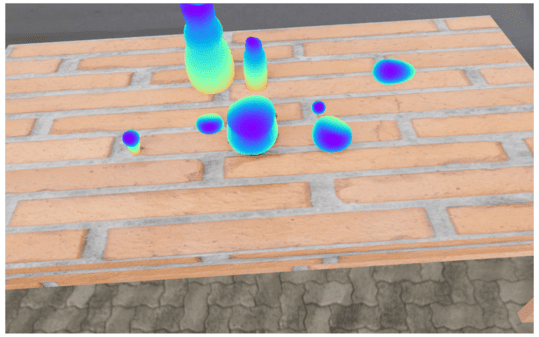}\end{tabular} \\

\begin{tabular}{l}\includegraphics[width=0.25\linewidth]{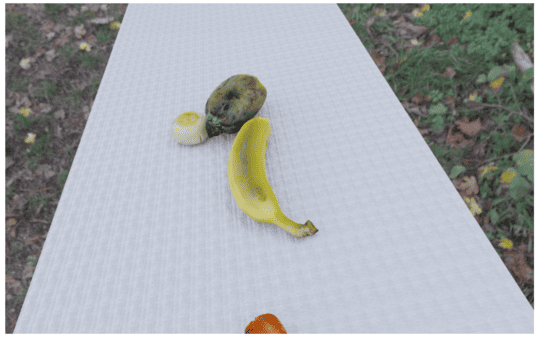}\end{tabular} &
\begin{tabular}{l}\includegraphics[width=0.25\linewidth]{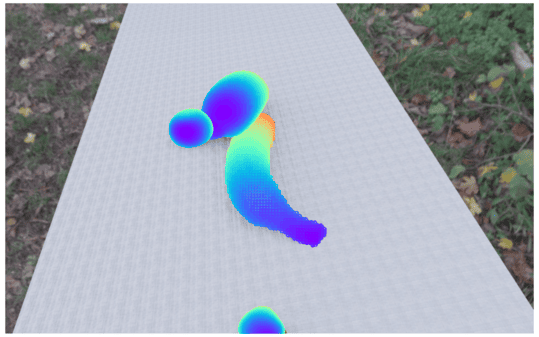}\end{tabular} &
\begin{tabular}{l}\includegraphics[width=0.25\linewidth]{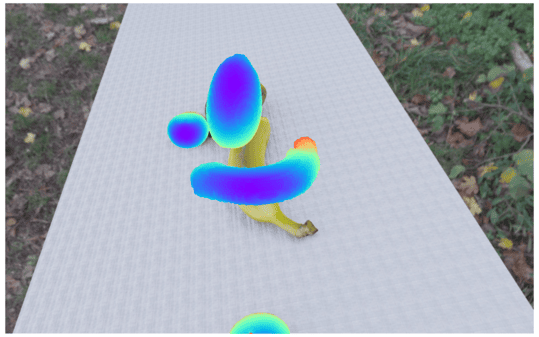}\end{tabular} &
\begin{tabular}{l}\includegraphics[width=0.25\linewidth]{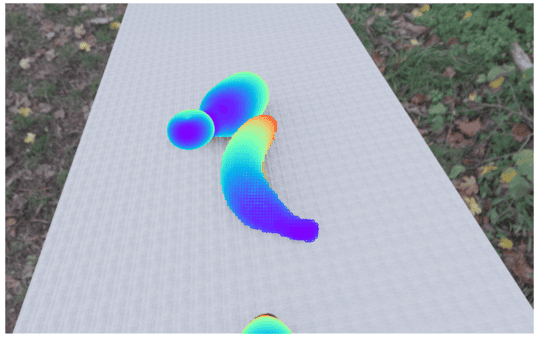}\end{tabular} \\

\begin{tabular}{l}\includegraphics[width=0.25\linewidth]{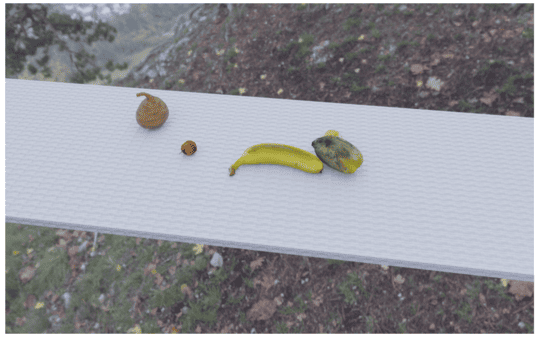}\end{tabular} &
\begin{tabular}{l}\includegraphics[width=0.25\linewidth]{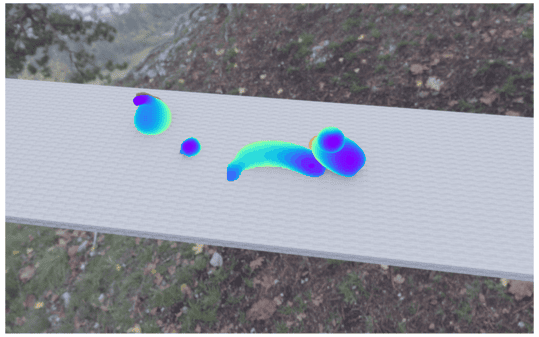}\end{tabular} &
\begin{tabular}{l}\includegraphics[width=0.25\linewidth]{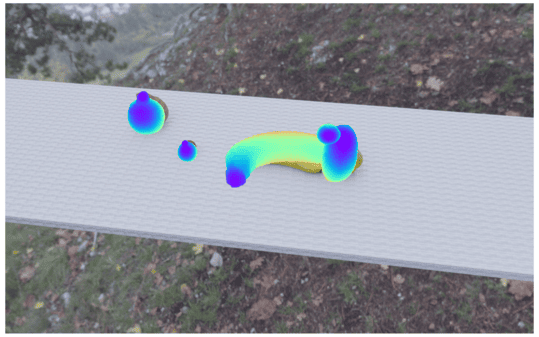}\end{tabular} &
\begin{tabular}{l}\includegraphics[width=0.25\linewidth]{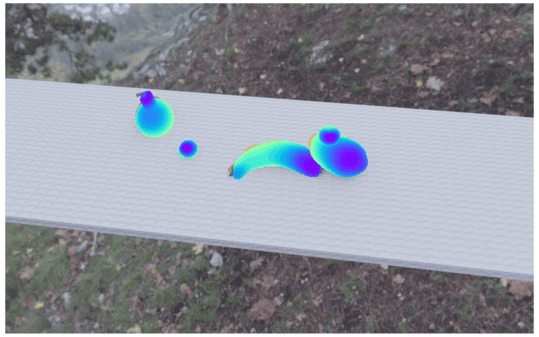}\end{tabular} \\

\begin{tabular}{l}\includegraphics[width=0.25\linewidth]{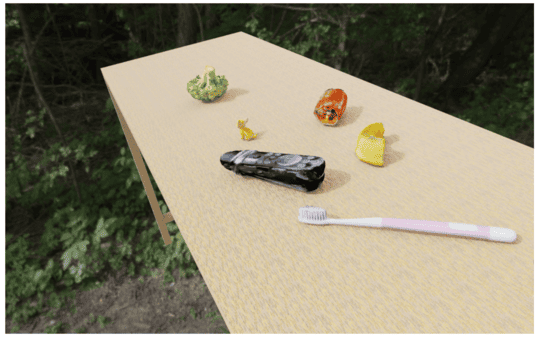}\end{tabular} &
\begin{tabular}{l}\includegraphics[width=0.25\linewidth]{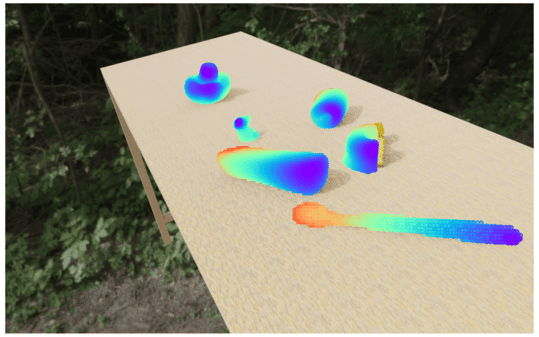}\end{tabular} &
\begin{tabular}{l}\includegraphics[width=0.25\linewidth]{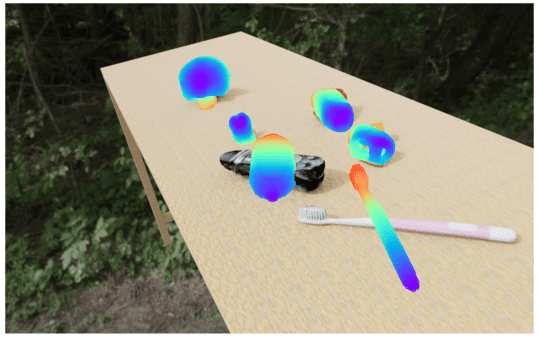}\end{tabular} &
\begin{tabular}{l}\includegraphics[width=0.25\linewidth]{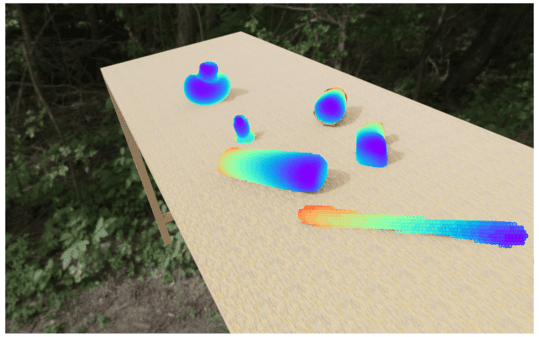}\end{tabular} \\

\end{tabular}}
\caption{\textbf{Qualitative results on LVS6D dataset.} 
    We present visualizations of Coders~\cite{zhang2024category} and UniPR on the LVS6D dataset. The results highlight our superior performance on the LVS6D dataset.
}
\label{fig:lvs_results_render}
\end{figure*}

We present qualitative results on LVS6D in \cref{fig:lvs_results_render}. 
Notably, the rotation predictions produced by UniPR demonstrate significant improvements compared to UniPR. 
In addition, UniPR achieves superior performance in mesh reconstruction.

\section{Detailed evaluation on public datasets}
\label{sec:sup-public}
\begin{figure*}[!t]
\setlength\tabcolsep{0.5 pt}
\centering
\scalebox{0.9}{
\begin{tabular}{ccccc}
GT &
\begin{tabular}{l}\includegraphics[height=2.5cm]{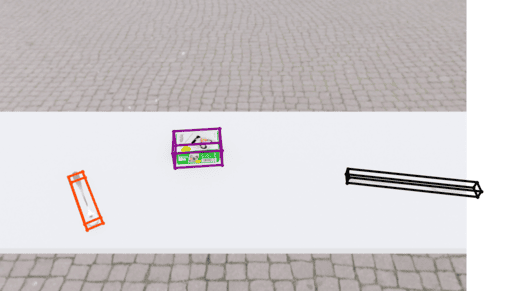}\end{tabular} &
\begin{tabular}{l}\includegraphics[height=2.5cm]{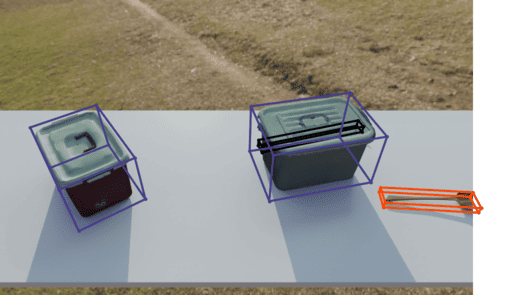}\end{tabular} &
\begin{tabular}{l}\includegraphics[height=2.5cm]{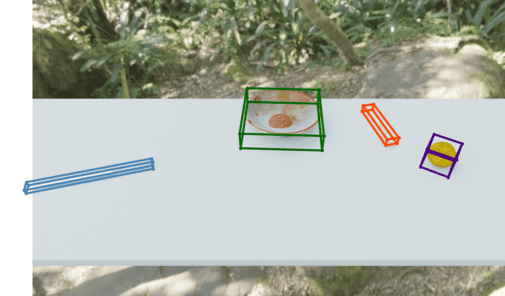}\end{tabular} &
\begin{tabular}{l}\includegraphics[height=2.5cm]{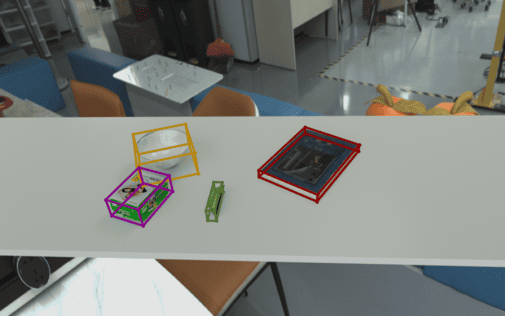}\end{tabular} \\

UniPR &
\begin{tabular}{l}\includegraphics[height=2.5cm]{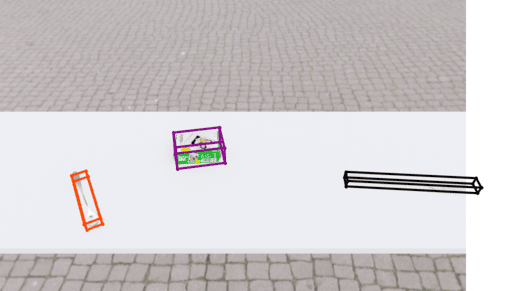}\end{tabular} &
\begin{tabular}{l}\includegraphics[height=2.5cm]{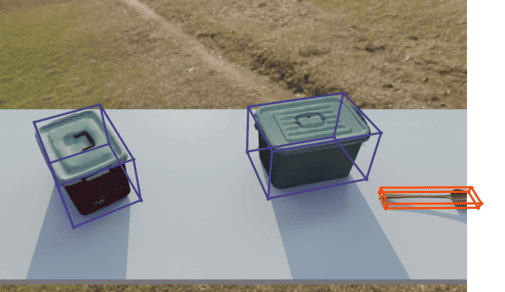}\end{tabular} &
\begin{tabular}{l}\includegraphics[height=2.5cm]{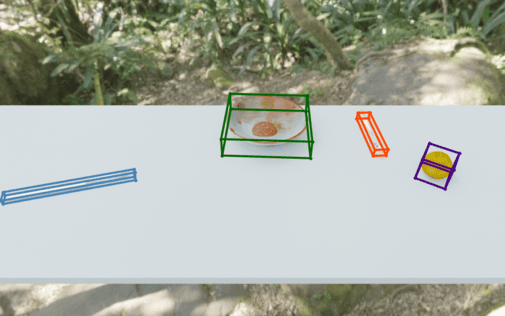}\end{tabular} &
\begin{tabular}{l}\includegraphics[height=2.5cm]{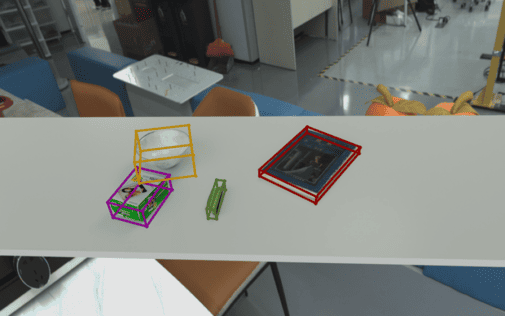}\end{tabular} \\

GT &
\begin{tabular}{l}\includegraphics[height=2.5cm]{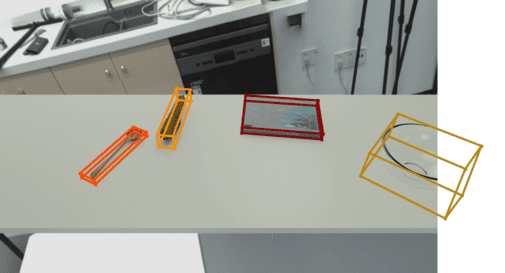}\end{tabular} &
\begin{tabular}{l}\includegraphics[height=2.5cm]{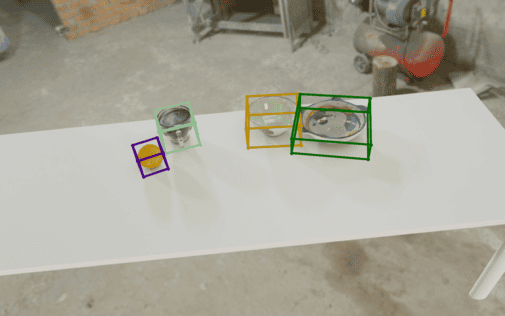}\end{tabular} &
\begin{tabular}{l}\includegraphics[height=2.5cm]{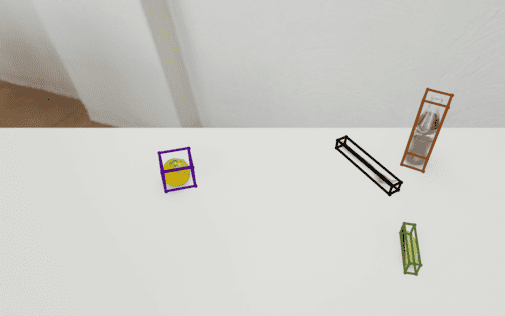}\end{tabular} &
\begin{tabular}{l}\includegraphics[height=2.5cm]{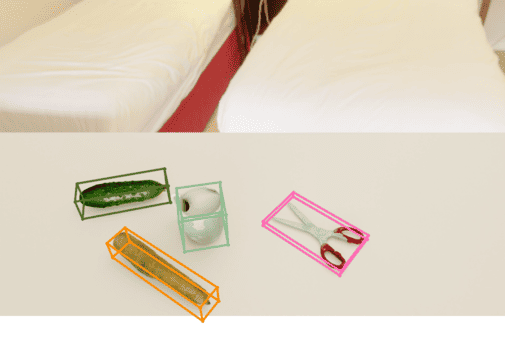}\end{tabular} \\

UniPR &
\begin{tabular}{l}\includegraphics[height=2.5cm]{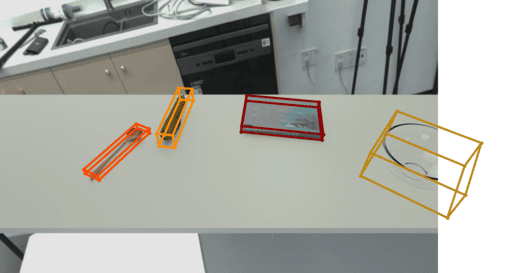}\end{tabular} &
\begin{tabular}{l}\includegraphics[height=2.5cm]{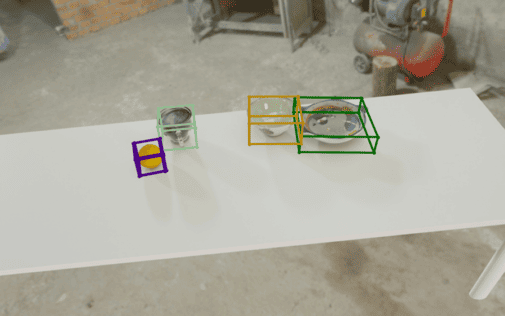}\end{tabular} &
\begin{tabular}{l}\includegraphics[height=2.5cm]{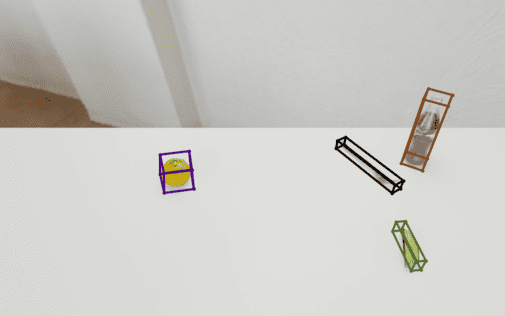}\end{tabular} &
\begin{tabular}{l}\includegraphics[height=2.5cm]{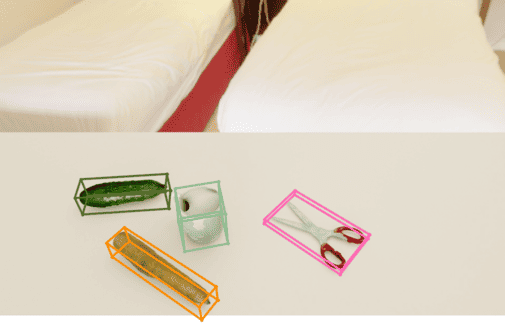}\end{tabular} \\

GT &
\begin{tabular}{l}\includegraphics[height=2.5cm]{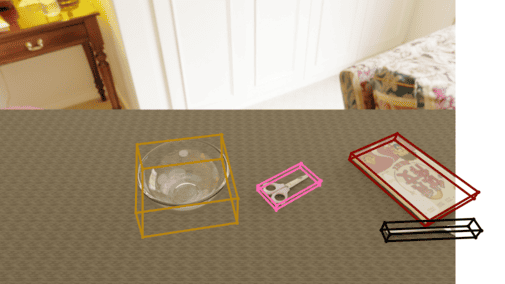}\end{tabular} &
\begin{tabular}{l}\includegraphics[height=2.5cm]{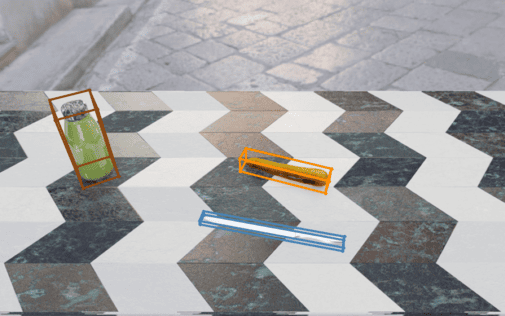}\end{tabular} &
\begin{tabular}{l}\includegraphics[height=2.5cm]{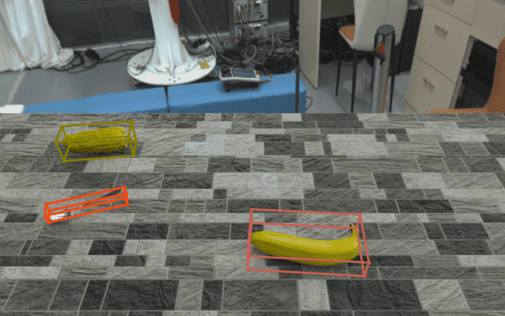}\end{tabular} &
\begin{tabular}{l}\includegraphics[height=2.5cm]{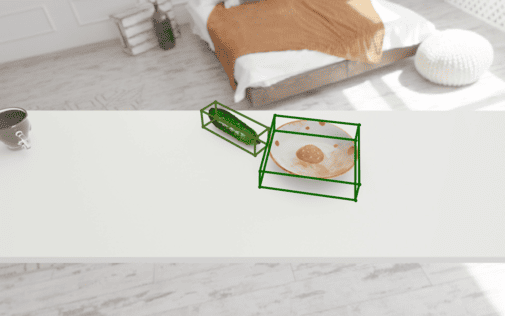}\end{tabular} \\

UniPR &
\begin{tabular}{l}\includegraphics[height=2.5cm]{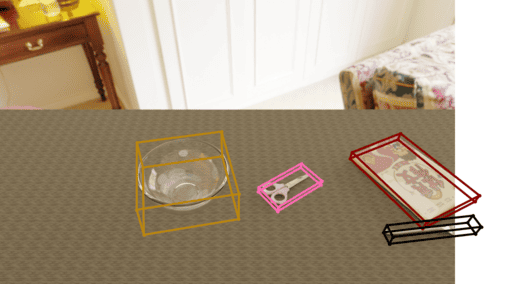}\end{tabular} &
\begin{tabular}{l}\includegraphics[height=2.5cm]{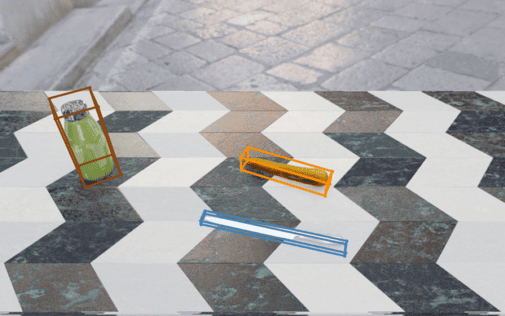}\end{tabular} &
\begin{tabular}{l}\includegraphics[height=2.5cm]{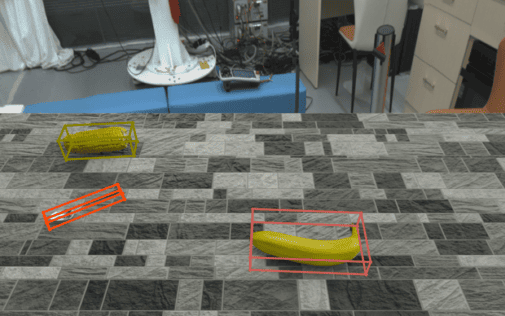}\end{tabular} &
\begin{tabular}{l}\includegraphics[height=2.5cm]{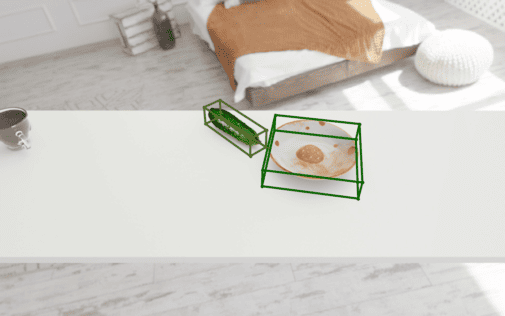}\end{tabular} \\

\end{tabular}}
\caption{\textbf{Visualization of StereoPose on SS3D dataset.} 
    Our proposed UniPR demonstrates consistently strong performance across various scenarios.
}
\label{fig:bbox}
\end{figure*}

We present a detailed evaluation of the 16 categories in the SS3D dataset which validates the effectiveness of our method. Additionally, we visualize the detection results in \cref{fig:bbox}.
Furthermore, \cref{tab:subsets_results} provides a detailed analysis of our method's performance on TOD.

\begin{table*}[!t]
\centering
\resizebox{0.95\linewidth}{!}{
\begin{tabular}{l|cccccccc}
\toprule
    \textbf{Category} & \textbf{Banana}  & \textbf{Bool}  & \textbf{Bottle} & \textbf{Bowl} & \textbf{Carrot}  & \textbf{Cork}  & \textbf{Cucumber}  & \textbf{Cup}   \\
\midrule
    Coders-$3D_{50}$          & 72 & 90 & 71 & 59 & 42 & 89 & 34 & 62 \\
\rowcolor{gray!10} \textbf{Ours}-$3D_{50}$  & \textbf{96}$\mathrm{\tiny{\hi{+24}}}$ & \textbf{97}$\mathrm{\tiny{\hi{+7}}}$ & \textbf{80}$\mathrm{\tiny{\hi{+9}}}$ & \textbf{98}$\mathrm{\tiny{\hi{+39}}}$ & \textbf{88}$\mathrm{\tiny{\hi{+46}}}$ & \textbf{97}$\mathrm{\tiny{\hi{+8}}}$ & \textbf{95}$\mathrm{\tiny{\hi{+61}}}$ & \textbf{98}$\mathrm{\tiny{\hi{+36}}}$ \\
    Coders-$3D_{75}$          & 26 & 51 & 16 & 20 & 11 & 45 & 9  & 16 \\
\rowcolor{gray!10} \textbf{Ours}-$3D_{75}$  & \textbf{60}$\mathrm{\tiny{\hi{+34}}}$ & \textbf{83}$\mathrm{\tiny{\hi{+32}}}$ & \textbf{54}$\mathrm{\tiny{\hi{+38}}}$ & \textbf{63}$\mathrm{\tiny{\hi{+37}}}$ & \textbf{59}$\mathrm{\tiny{\hi{+48}}}$ & \textbf{69}$\mathrm{\tiny{\hi{+24}}}$ & \textbf{57}$\mathrm{\tiny{\hi{+48}}}$ & \textbf{59}$\mathrm{\tiny{\hi{+43}}}$ \\
    Coders-$5^{\circ} 5\mathrm{cm}$    & 47 & 65 & \textbf{80} & 58 & 51 & 74 & 38 & 72 \\
\rowcolor{gray!10} \textbf{Ours}-$5^{\circ} 5\mathrm{cm}$ & \textbf{75}$\mathrm{\tiny{\hi{+28}}}$ & \textbf{87}$\mathrm{\tiny{\hi{+22}}}$ & 75$\mathrm{\tiny{\hi{-5}}}$ & \textbf{96}$\mathrm{\tiny{\hi{+38}}}$ & \textbf{88}$\mathrm{\tiny{\hi{+37}}}$ & \textbf{90}$\mathrm{\tiny{\hi{+16}}}$ & \textbf{84}$\mathrm{\tiny{\hi{+46}}}$ & \textbf{96}$\mathrm{\tiny{\hi{+24}}}$  \\
    Coders-$5^{\circ} 2\mathrm{cm}$    & 26 & 43 & 33 & 23 & 22 & 48 & 20 & 33 \\
\rowcolor{gray!10} \textbf{Ours}-$5^{\circ} 2\mathrm{cm}$   & \textbf{34}$\mathrm{\tiny{\hi{+8}}}$ & \textbf{48}$\mathrm{\tiny{\hi{+5}}}$ & \textbf{39}$\mathrm{\tiny{\hi{+6}}}$ & \textbf{63}$\mathrm{\tiny{\hi{+40}}}$ & \textbf{53}$\mathrm{\tiny{\hi{+31}}}$ & \textbf{49}$\mathrm{\tiny{\hi{+1}}}$ & \textbf{40}$\mathrm{\tiny{\hi{+20}}}$ & \textbf{61}$\mathrm{\tiny{\hi{+28}}}$  \\
\midrule
    \textbf{Category}& \textbf{Dish}  & \textbf{Fork}  & \textbf{Knife}  & \textbf{LargeBox}  & \textbf{Orange}  & \textbf{SmallBox}  & \textbf{Scissors}  & \textbf{Spoon}  \\
\midrule
    Coders-$3D_{50}$          & 88 & 54 & \textbf{71} & 52 & 35 & 66 & 48 & 42 \\
\rowcolor{gray!10} \textbf{Ours}-$3D_{50}$  & \textbf{95}$\mathrm{\tiny{\hi{+7}}}$ & \textbf{80}$\mathrm{\tiny{\hi{+26}}}$ & 68$\mathrm{\tiny{\hi{-3}}}$ & \textbf{90}$\mathrm{\tiny{\hi{+38}}}$ & \textbf{95}$\mathrm{\tiny{\hi{+60}}}$ & \textbf{94}$\mathrm{\tiny{\hi{+28}}}$ & \textbf{94}$\mathrm{\tiny{\hi{+46}}}$ & \textbf{55}$\mathrm{\tiny{\hi{+13}}}$ \\
    Coders-$3D_{75}$          & 37 & 15 & 12 & 9  & 6  & 32 & 18 & 10 \\
\rowcolor{gray!10} \textbf{Ours}-$3D_{75}$  & \textbf{56}$\mathrm{\tiny{\hi{+19}}}$ & \textbf{36}$\mathrm{\tiny{\hi{+21}}}$ & \textbf{45}$\mathrm{\tiny{\hi{+33}}}$ & \textbf{69}$\mathrm{\tiny{\hi{+60}}}$ & \textbf{41}$\mathrm{\tiny{\hi{+35}}}$ & \textbf{55}$\mathrm{\tiny{\hi{+23}}}$ & \textbf{50}$\mathrm{\tiny{\hi{+32}}}$ & \textbf{29}$\mathrm{\tiny{\hi{+19}}}$ \\
    Coders-$5^{\circ} 5\mathrm{cm}$    & 64 & 58 & 61 & 16 & 56 & 32 & 51 & 33 \\
\rowcolor{gray!10} \textbf{Ours}-$5^{\circ} 5\mathrm{cm}$  & \textbf{79}$\mathrm{\tiny{\hi{+15}}}$ & \textbf{70}$\mathrm{\tiny{\hi{+12}}}$ & \textbf{62}$\mathrm{\tiny{\hi{+1}}}$ & \textbf{69}$\mathrm{\tiny{\hi{+53}}}$ & \textbf{95}$\mathrm{\tiny{\hi{+39}}}$ & \textbf{57}$\mathrm{\tiny{\hi{+25}}}$ & \textbf{77}$\mathrm{\tiny{\hi{+26}}}$ & \textbf{44}$\mathrm{\tiny{\hi{+11}}}$ \\
    Coders-$5^{\circ} 2\mathrm{cm}$    & 20 & 33 & \textbf{33} & 5  & 25 & 24 & 29 & \textbf{20} \\
\rowcolor{gray!10} \textbf{Ours}-$5^{\circ} 2\mathrm{cm}$   & \textbf{37}$\mathrm{\tiny{\hi{+17}}}$ & \textbf{41}$\mathrm{\tiny{\hi{+8}}}$ & 31$\mathrm{\tiny{\hi{-2}}}$ & \textbf{26}$\mathrm{\tiny{\hi{+21}}}$ & \textbf{52}$\mathrm{\tiny{\hi{+27}}}$ & \textbf{26}$\mathrm{\tiny{\hi{+2}}}$ & \textbf{33}$\mathrm{\tiny{\hi{+4}}}$ & 19$\mathrm{\tiny{\hi{-1}}}$ \\
\bottomrule
\end{tabular}
}
\caption{\textbf{Detailed Results on the SS3D Test Dataset.}
We evaluate UniPR on the SS3D test dataset with 16 categories of unseen objects. The first row corresponds to categories in reference to SS3D.
Our method can manage objects across all 16 categories, encompassing various sizes, shapes and materials, demonstrating the generalization capability of our stereo framework. The gray rows in the table indicate results from our method and the numbers in the table represent accuracy percentages.
}
\label{tab:subsets_results}
\end{table*}

